%% file: main.tex
\documentclass{article}

\PassOptionsToPackage{numbers}{natbib}
\usepackage[preprint]{neurips_2026}
\makeatletter
\renewcommand{\@notice}{}
\makeatother

\usepackage{xspace}
\usepackage{url}
\usepackage{acronym}
\usepackage{booktabs}
\usepackage{enumitem}
\usepackage{amsmath}
\usepackage{graphicx}
\usepackage{pifont}
\usepackage{xcolor}
\definecolor{priorlabs}{HTML}{101075}
\usepackage{fontawesome5}
\usepackage{makecell}
\usepackage{hyperref}
\usepackage[capitalise]{cleveref}

\input{macros}

\title{Advancing Open and Reproducible Relational Learning:
\RelArena, \TabPFNRel and \RPI}
\author{%
\parbox{\textwidth}{%
\begin{tabular}{@{}l@{\hspace{0.8em}}l@{}}
  {\itshape\footnotesize Core Contributors:} &
    \textbf{Adrian Hayler}$^{1}$ \quad \textbf{Klemens Fl\"oge}$^{1}$ \quad \textbf{Alan Arazi}$^{1}$ \\[3pt]
  {\itshape\footnotesize External Collaborators:} &
    \textbf{Rishabh Ranjan}$^{2}$ \quad \textbf{Jure Leskovec}$^{2,3}$ \\[3pt]
  {\itshape\footnotesize Research \& Project Leads:} &
    \textbf{Lennart Purucker}$^{1,4}$ \quad \textbf{Frank Hutter}$^{1,4,5}$ \quad \textbf{Noah Hollmann}$^{1}$ \\
\end{tabular}}%
\\
\rule{0pt}{12pt}\textbf{and the Prior Labs Team}$^{1}$
\\[12pt]
    \normalsize $^{1}$Prior Labs \quad $^{2}$Stanford University
    \quad
   $^{3}$NVIDIA
       \\
   \quad
   $^{4}$University of Freiburg
   \quad
  $^{5}$ELLIS Institute T\"ubingen \\[9pt]
  \raisebox{-0.35\height}{\includegraphics[height=22pt]{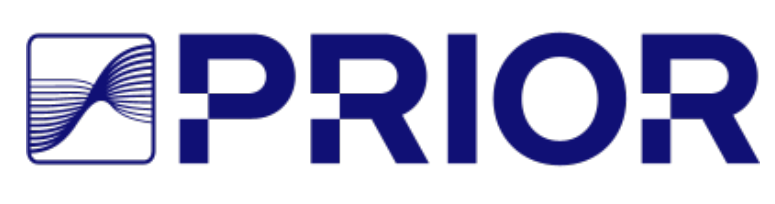}}%
  \hspace{22pt}%
  \raisebox{-0.35\height}{\includegraphics[height=26pt]{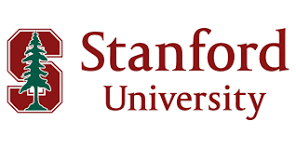}}
   \hspace{18pt}%
  \raisebox{-0.35\height}{\includegraphics[height=26pt]{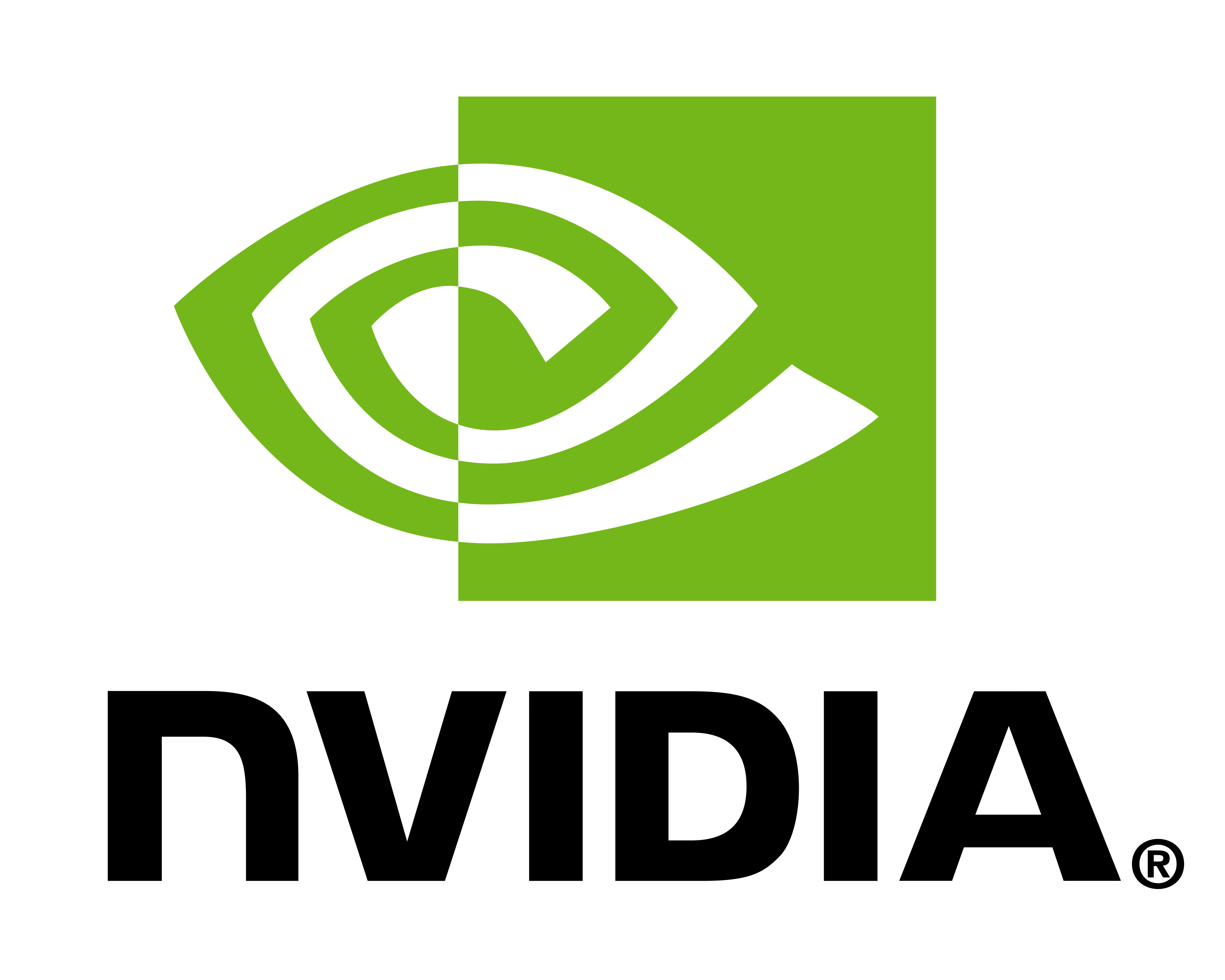}}%
  \\
  \\
  \hspace{2pt}%
  \raisebox{-0.35\height}{\includegraphics[height=13pt]{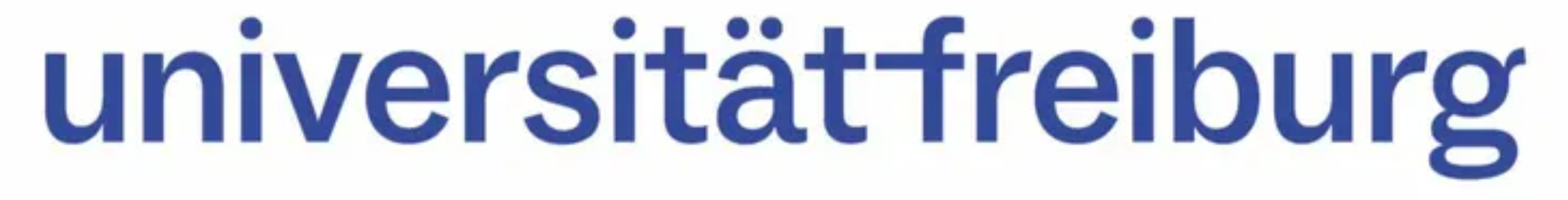}}%
  \hspace{10pt}%
  \raisebox{-0.35\height}{\includegraphics[height=16pt]{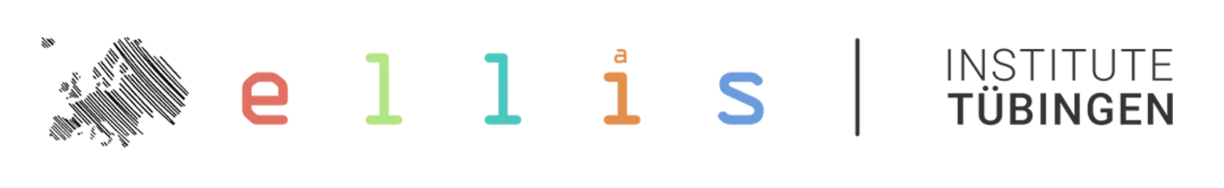}}%
}
\date{}

\begin{document}

\maketitle

\begin{abstract}

This first release of Prior Labs in \rellearning shows our continued commitment to open science. We open-source three pieces of software that we expect to accelerate research in the field towards meaningful real-world impact.
We aim to steer further development based on feedback from, and in collaboration with, the community. Given the early stage of development, our $\alpha$-release targets researchers and early-adopting practitioners.

\paragraph{\RelArena.} Over the past years, a variety of datasets and tasks for \rellearning have emerged~\cite{relbenchv1, 4dbinfer, redelex, relbenchv2}, but the community has not converged on a reliable, reproducible way to compare different methods on these tasks. Our $\alpha$-release, \RelArena, provides a unified framework for running and comparing baselines on RelBench v1~\cite{relbenchv1} by standardizing data loading, evaluation protocols, tuning regimes, and support for systems with custom tuning, inspired by established tabular benchmarks such as TabArena~\cite{tabarena}. We plan to work with the research community to further develop \RelArena into a catalyst for progress in the \rellearning community. 

\paragraph{\TabPFNRel.} We release the initial version of \TabPFNRel, a purpose-built relational harness for \TabPFNThree. Currently ranked first among models on \RelArena, \TabPFNRel makes key improvements upon RDBLearn~\cite{rdblearn}. Beyond its ranking, \TabPFNRel serves as a strong baseline, adding to the growing evidence that flattening a relational database into a single table remains competitive with specialized relational architectures on real-world tasks~\cite{rdblearn, rdblearncompanion, relagent, paramfreeviable}.

\paragraph{\RPI.} To facilitate adoption of \rellearning methods in research and industry, we release an initial $\alpha$-version of our \textbf{R}elational \textbf{P}redictive \textbf{I}nterface, \RPI, an open-source, model-agnostic interface that enables early adopters to easily define problems on new databases and apply any model implemented in \RelArena, including \TabPFNRel, to these problems.

\end{abstract}

% Pointer to the released code, below the abstract. The link border is turned
% off locally because the colour already marks it as a link.
\begin{center}
\begingroup\hypersetup{pdfborder={0 0 0}}%
\href{https://github.com/PriorLabs/relarena}%
  {\color{priorlabs}\faGithub~\texttt{https://github.com/PriorLabs/relarena}}%
\endgroup
\end{center}

\newpage

\section{The Current State of Relational Learning}
\label{sec:problems}

Relational learning has attracted increased interest and research output in recent years. Nevertheless, we believe that progress in the field has been slowed and real-world impact hindered by the following interlocking factors.

\begin{itemize}[leftmargin=2em]
    \item \textbf{Methods are not reproducible.} Several of the strongest-reported methods cannot be independently reproduced, as training scripts are unavailable and tuning procedures are undisclosed.
    \item \textbf{Methods are not comparable.} Reported results are often obtained under differing evaluation protocols and tuning budgets, making end-to-end performance comparisons inherently confounded.
    As a result, apparent differences in \emph{model} performance may reflect differences in the surrounding \emph{system} rather than differences between the models themselves (e.g., tuning, validation protocol).
    The community lacks standardized guidance for comparing models and systems under controlled evaluation regimes that allow reliable comparisons. 
    \item \textbf{No shared aggregation standard.} The community lacks a common convention for aggregating results across tasks, making overall performance claims difficult to compare and interpret.
    \item \textbf{Lack of pragmatic baselines.} Feature-engineering baselines prevalent in industry are underrepresented in academic evaluations and are frequently implemented without sufficient care.
    \item \textbf{No path from paper to practice.} Published methods lack accessible interfaces that would allow practitioners to apply them to their own databases.
\end{itemize}

\Cref{tab:issues_methods} summarizes, for the methods discussed throughout this section, which of these issues apply.

\newcommand{\defaultcolwidth}{1.38cm}

\begin{table}[h]
\centering
\footnotesize
\setlength{\tabcolsep}{2pt}
\setlength{\aboverulesep}{0.2ex}
\setlength{\belowrulesep}{0.2ex}
\setlength{\belowcaptionskip}{6pt}

\caption{\textbf{Reproducibility and comparability of methods reporting RelBench v1 results.} State at the time of submission.
We summarize code availability, tuning practices, evaluation regimes, aggregation methods, task coverage, and packaging, which we discuss in detail in \Cref{sec:problems}. Search-space sizes are taken from documented procedures or inferred from released per-task configurations where the tuning procedure is undisclosed. \RelArena uses method-specific trial budgets chosen to approximately balance tuning compute (see Appendix~\ref{app:tuning}).}
\label{tab:issues_methods}

\makebox[\textwidth][c]{%
\begin{tabular}{@{}
    >{\raggedright\arraybackslash}m{1.6cm}
    >{\centering\arraybackslash}m{1.6cm}
    >{\centering\arraybackslash}m{1.1cm}
    >{\centering\arraybackslash}m{\defaultcolwidth}
    >{\centering\arraybackslash}m{\defaultcolwidth}
    >{\centering\arraybackslash}m{\defaultcolwidth}
    >{\centering\arraybackslash}m{1.7cm}
    >{\centering\arraybackslash}m{1.2cm}
    >{\centering\arraybackslash}m{\defaultcolwidth}
@{}}

\toprule
Method
& Code for all reported tasks
& Per-task tuning
& Tuning documented
& Search space size
& RelBench eval.\ regime
& Regression aggregate
& \#Tasks reported
& Installable package \\
\midrule

GraphSAGE
& \cmark & \xmark & -- & -- & \cmark
& mean MAE & 21 & \xmark \\

RelGNN
& \xmark & \cmark & \xmark & up to 25k & \cmark
& -- & 21 & \xmark \\

RelGT
& \cmark & \cmark & \cmark & 9 & \cmark
& -- & 21 & \xmark \\

KumoRFM
& \xmark & \cmark & \xmark & up to 10 & \xmark
& MAE / RDL & 21 & \xmark \\

KumoRFM-2
& \xmark & \cmark & \xmark & up to 384 & \xmark
& MAE / LGBM & 21 & \xmark \\

RT
& \cmark & \xmark & -- & -- & \xmark
& mean $R^2$ & 18 & \xmark \\

PluRel
& \cmark & \xmark & -- & -- & \xmark
& mean $R^2$ & 18 & \xmark \\

RT-J
& \cmark & \cmark & \cmark & 32 & \xmark
& $\sigma$-norm. MAE & 21 & \xmark \\

RDBLearn
& \cmark & \cmark & \cmark & 9 & \cmark
& MAE / AG & 16 & \cmark \\

\midrule

\RelArena
& \cmark & \cmark & \cmark & $\sim$matched & \cmark
& Elo (+others) & 21 & \cmark \\

\bottomrule
\end{tabular}%
}

\end{table}

\subsection{Methods are not reproducible}
\label{sec:problems:not_reproducible}

\paragraph{Impossible to reproduce.} While many \rellearning methods release all code required for reproduction, the level of reproducibility varies across methods. In some cases, released implementations provide dataset-specific model checkpoints but not the corresponding training scripts~\cite{relgnn}, despite community interest in obtaining the training code~\cite{relgnntrainingcodeissue}. In other cases, foundation models are made available through an API, but the provided scripts do not cover all tasks reported in the paper~\cite{kumorfmv2}. These differences in availability make it difficult to verify results under a common experimental setup and to perform controlled comparisons across methods.

\paragraph{Opaque tuning regimes.} Similarly, even when results can be reproduced via an API or model checkpoint, performance is often contingent on dataset-specific hyperparameters, which are often selected from grids spanning hundreds or thousands of configurations. Large configuration spaces are not problematic in themselves, but without a documented tuning regime it is unclear how extensively they were explored, how final configurations were selected and if the tuning regime is transferable to real-world settings. For KumoRFM-2, the released per-task configurations vary along five dimensions whose observed values alone induce a grid of 384 combinations~\cite{kumorfmbenchmarkscripts}; the released RelGNN checkpoints vary along nine dimensions whose observed values induce a grid of over 25,000 combinations~\cite{relgnntuningissue}.
Neither the publications nor the underlying codebases explain how these hyperparameters were obtained~\cite{relgnn, kumorfmv2}, and public requests for clarification remained unresolved at the time of this release~\cite{relgnntuningissue, kumorfmhparamsissue}. Without this information, it is difficult to assess how much of the reported performance reflects the method itself rather than method-agnostic optimizations.

\subsection{Methods are not comparable}
\label{sec:problems:not_comparable}

Unfortunately, it has become common practice in the \rellearning community to copy self-reported baseline results for method comparisons~\cite{relgnn, relgt, kumorfmv1, kumorfmv2, rdblearn, rdblearncompanion, relagent, rt-j, paramfreeviable}, a practice we also followed in pursuit of ``comparability'' in a recent publication~\cite{tabpfn_3_model_report}. This practice may partly stem from the reasons outlined above, which sometimes force authors to copy reported baseline results if they want to, or have to, compare against a specific method. Still, even assuming baseline results are reproducible, the reported baselines are, with few exceptions, not comparable, as we outline below.

\paragraph{Differences in evaluation regimes.} 
Defining a task in \rellearning is complex and not standardized. As a result, subtle differences in evaluation regimes have emerged over time across method releases. These subtle differences can easily lead to significant downstream performance differences, undermining the scientific integrity of the method comparison.
\\
To illustrate, consider the entity-level forecasting tasks in RelBench~\cite{relbenchv1}, which have been widely used by the community.
The entity-level forecasting tasks in RelBench freeze the database at a fixed, pre-defined test cut-off, independent of the specific timestamp of a test entity.
In particular, test entities do not have access to the label of other test entities with earlier timestamps. This evaluation regime is defined in RelBench and enforced by the provided data-loading logic. However, not every method directly uses RelBench's data loading, which enables methods to utilize the additional information ingested into the database between the fixed test cut-off and the entity's timestamp. While most methods follow the former evaluation regime, we observed that KumoRFM-2~\cite{kumorfmv2} (and likely its predecessor, KumoRFM~\cite{kumorfmv1}) as well as methods derived from the relational transformer architecture~\cite{rt, plurel, rt-j} follow the latter regime.
While this difference materially affects only one of the seven RelBench v1 databases, it still has a very meaningful impact on aggregate statistics. On every database other than \texttt{rel-f1}, all test entities share a single timestamp equal to the test cut-off, so both regimes expose the same database state. On the \texttt{rel-f1} tasks, however, test timestamps span three to six years, and methods following the latter regime gain access to years of additional signal: evaluating the same model under RelBench's regime instead reduces its results by over ten ROC AUC points on the two classification tasks and increases its MAE by over 40\% on the regression task~\cite{tabpfn_3_model_report, kumorfmv2}.
Both evaluation regimes are valid, but the difference in additional context makes them incomparable. In a fair comparison, all methods would use the same evaluation regime.

\paragraph{Differences in data states.}
The RelBench maintainers actively address issues in databases and tasks as they are discovered. Defects of this kind are to be expected in any benchmark of this scale. However, as a community, we need to account for these fixes in method comparisons; they change the data state and may invalidate older baseline results, which, in turn, need to be recomputed.
\\
To illustrate, consider the temporal-leakage fix released in early 2026~\cite{relbench_leakage_fix}, which regenerated the labels for the \texttt{rel-event/user-ignore} task. When affected methods are re-evaluated on the corrected task, their ROC AUC scores are consistently two to six points lower than the originally reported values~\cite{relbenchv1, relgnn, relgt} (cf.\ our re-runs in Appendix~\ref{app:raw}). Methods evaluated against different versions of the RelBench package may therefore be evaluated on materially different data states, rendering direct comparisons of their self-reported results impossible.

\paragraph{Differences in tuning.} 
Across many subfields of machine learning, model performance is known to be sensitive to dataset-specific hyperparameters~\cite{gnnpitfalls, welltunednets,hposurvey,rlmatters, tabarena}. 
Comparing a tuned to an untuned method confounds the comparison. The tuning alone could explain the performance difference. 
The same holds when comparing a method tuned under one regime with one tuned under a different regime (e.g., random search vs. Bayesian optimization).
Hence, it is vital to standardize the tuning regime across models to eliminate tuning as a confounding factor. 
At the same time, it is vital not to suppress research ingenuity and to enable benchmarking of novel tuning regimes. 
Unfortunately, it remains acceptable in the \rellearning literature for authors to demonstrate the merit of their isolated contribution with comparisons confounded by the tuning regimes. 
\\
Even popular methods, such as RelGNN~\cite{relgnn} and RelGT~\cite{relgt}, follow this pattern. Both build upon and compare against the relational deep learning (RDL) baseline~\cite{rdl_fey} by modifying the underlying model architecture. Additionally, both methods apply tuning to improve model performance. Yet the authors compare only against results reported in~\citet{relbenchv1}, which were obtained without tuning.
``RDL'' could have been run with the same generic search space as RelGNN or RelGT, as they use the same underlying data pipeline; or RelGNN and RelGT could have positioned their tuning as an integral part of their contributions. 
Without fair, rigorous method comparisons, it is impossible to determine whether improvements over baselines come from a novel tuning regime or methodological improvements; this misleads researchers and confuses practitioners.

\subsection{No shared aggregation standard}
\label{sec:problems:aggregation}

Even with a fixed evaluation regime and a standardized tuning budget, per-dataset results still have to be aggregated before general conclusions can be drawn, and here the community has no shared convention. Work that uses RelBench reports the same per-task metrics (ROC AUC for classification, MAE for regression), but summarizes them in incompatible ways. While classification results are aggregated consistently (as the unweighted mean of raw ROC AUC scores~\cite{relbenchv1, kumorfmv2, rt}), for regression, the aggregates use MAE normalized by the performance of different reference models before averaging: LightGBM~\cite{kumorfmv2, relagent}, the RDL baseline~\cite{kumorfmv1}, a non-relational AutoGluon~\cite{rdblearn}, or RelGT~\cite{rdblearncompanion}. Other approaches include means of raw MAE across tasks whose scales differ by orders of magnitude~\cite{relbenchv1}, a geometric mean of raw MAE~\cite{paramfreeviable}, standardization by the target's standard deviation~\cite{rt-j}, and replacing the metric with $R^2$ altogether~\cite{rt, plurel}. Further summaries include average relative improvement over a baseline~\cite{relbenchv1, kumorfmv1}, win counts~\cite{relgnn, relgt}, and mean ranks~\cite{kumorfmv2, rdblearn, relagent, paramfreeviable}. Several papers also compute their aggregates over subsets of the available tasks~\cite{rt, plurel, rdblearn, rdblearncompanion, paramfreeviable}, dropping between three and six of the 21 tasks for reasons ranging from suspected data leakage (see also \cref{sec:problems:not_comparable}) to no stated rationale.
\\
These summaries are not interchangeable and carry different interpretations. Averaging raw ROC AUC values gives higher weight to tasks with larger absolute differences, while down-weighting tasks with meaningful differences that are small in absolute value. A reference-model-normalized MAE has a different interpretation because it weights absolute differences relative to the performance of the reference model (\eg, across two datasets, it rewards the same absolute difference less when the reference model performs worse on one). Summaries over an incomplete result table can additionally favor methods that report only subsets of tasks on which they perform well.
\\
Because each paper uses its own summary, aggregate numbers are rarely comparable across papers, leaving the field without a common quantity against which method development can be measured. This is less of a problem in adjacent communities, which have converged on shared approaches to summarizing per-dataset results: language model comparisons use Elo ratings derived from pairwise outcomes~\cite{chiang2024chatbot}, while the tabular literature also reports normalized scores and mean ranks, along with significance tests across datasets~\cite{tabarena, demvsar2006statistical}. A comparable convention for \rellearning would make general trends easier to interpret and give the community a common target to optimize for.

\subsection{Lack of pragmatic baselines}

Researchers must compare against pragmatic baselines that industry practitioners currently use to solve prediction tasks on \acp{rdb}. Such methods typically convert the relational learning task into a tabular one by flattening the \ac{rdb} into a table~\cite{rdl_fey, relbenchv1, redelex}. On such a table, traditional tabular learning methods can be applied to produce predictions. There are many ways to convert an \ac{rdb} into a flat table; these generally fall into automated and manual feature engineering.

\paragraph{Manual feature engineering.} An important contribution of RelBench v1~\cite{relbenchv1} is the inclusion of a ``data scientist'' baseline that combines manual feature engineering with LightGBM~\cite{lightgbm}. This baseline is particularly valuable because it reflects how industry practitioners would approach these tasks. At the same time, it should be viewed as a standardized, reproducible reference rather than an estimate of the best performance achievable through manual feature engineering. In practice, data scientists typically refine their feature engineering based on validation set performance. RelBench's division of the data scientist workflow into fixed sequential steps cannot capture this iterative feedback loop~\cite{relbenchv1}. Recent work suggests that iterative feature refinement can yield substantial performance improvements~\cite{relagent}.

\paragraph{Automated feature engineering.} While automated feature engineering approaches~\cite{featuretools, onebm, getml} have been popular in industry, they have been ignored as possible baselines for years. Only recently have they been evaluated as baselines in the \rellearning literature, where they have achieved competitive performance from the outset~\cite{rdblearn, rdblearncompanion}. Before this, baselines that lift tabular learners to \rellearning via automated feature engineering provided the model with only a handful of raw columns from the entity's own table~\cite{relbenchv1, relbenchv2}, making for a weak and unrepresentative baseline. While these baselines significantly underestimate the power of tabular learning for \rellearning tasks, they are prominently used to demonstrate the advantages of GNN-based approaches over flattening with tabular learning~\cite{relbenchv1, relbenchv2}. 
\\
In extreme cases, tabular baseline implementations contain obvious errors. For the autocomplete tasks of RelBench v2, we noticed that the reported LightGBM baseline has a 50\% ROC-AUC and negative $R^2$. 
After a quick investigation, we found that the LightGBM baseline receives only \texttt{NaN} inputs at test time, resulting in 50\% ROC-AUC and negative $R^2$. This example illustrates that baselines require the same implementation care and validation as newly proposed methods.

\subsection{No path from paper to practice}

Despite these shortcomings, many exciting developments in \rellearning over the past few years could benefit industry practitioners. At the same time, transferring these methods from research benchmarks to real-world prediction problems remains unnecessarily difficult.
\\
Research methods are typically developed around benchmark tasks represented through task tables~\cite{rdl_fey, relbenchv1, 4dbinfer, relbenchv2}. RelBench v1 provides tutorials for extending the benchmark to custom databases and prediction tasks. However, doing so still requires users to write task-specific Python code and interact with RelBench's dataset and task abstractions. This suits researchers extending the benchmark, but it does not provide a dedicated declarative interface through which practitioners can specify a prediction problem on an existing relational database.
\\
To the best of our knowledge, the only attempt to establish a dedicated query language for this purpose has been the Predictive Query Language (PQL)~\cite{pql}, whose implementations remain proprietary. Moreover, PQL's published grammar cannot express some RelBench tasks, whose labels require multi-hop aggregations (\texttt{rel-avito/user-clicks} and \texttt{rel-trial/study-outcome}).
\\
Application of \rellearning methods is further hindered by the absence of a unified model interface comparable to scikit-learn~\cite{scikit_learn} for tabular learning or HuggingFace Transformers~\cite{transformers} for language models. Switching between methods therefore requires custom integration logic for each model and task. Compounding this, currently most \rellearning methods are not distributed as installable packages, requiring practitioners to work directly with the individual codebases.

\section{Towards Open, Reproducible, and Accessible Relational Learning}
\label{sec:contribution}

Our first open-source release in \rellearning is a first step towards addressing the issues raised in the previous section: \RelArena targets reproducibility and comparability, \TabPFNRel addresses the lack of pragmatic baselines, and \RPI makes \rellearning research accessible to industry practitioners. While all released artifacts are still in early development, we believe they could already benefit the community. 
We ask the community for feedback and contributions to accelerate the development of an open, reproducible, and accessible \rellearning ecosystem.

\subsection{\RelArena}
\label{sec:contribution:relarena}

Inspired by the success of TabArena~\cite{tabarena} and its recent extension BeyondArena~\cite{purucker2026beyond} in enabling fair and reproducible benchmarking for tabular learning, we introduce \RelArena, with the goal of bringing the same standard to \rellearning. \RelArena redistributes no data, retrieving all databases at runtime through RelBench under their respective licenses (see Appendix~\ref{app:provenance}). For its initial $\alpha$-release, \RelArena focuses on RelBench v1's entity-level forecasting tasks. We choose this scope for two reasons. First, entity-level forecasting is by far the most widely used task type in the current \rellearning literature~\cite{relgt, relgnn, rt, rt-j, plurel, kumorfmv1, kumorfmv2, rdblearn, akhter2026fair}, making it the natural starting point for establishing a common comparison framework. Second, methods developed for one relational task type do not necessarily transfer directly to others, making it difficult to define a unified benchmark across task types at this stage. We therefore view the current task set as a pragmatic starting point rather than a definitive benchmark; broader questions around task and dataset curation remain open, as discussed in \cref{sec:open_issues}. We detail our main contributions below.

\paragraph{Reproducibility of methods.} We ensure the reproducibility of methods by re-running all methods reported on \RelArena through a unified model API. We have invested significant effort to align model implementations with our benchmark API, fix bugs, and, if necessary, reconstruct training scripts that were not provided by the authors. We have spent hundreds of GPU hours producing baseline results that the community can trust, while also providing an accessible way for anyone to reproduce our reported results.

\paragraph{Tuning regimes for models and systems.}
We enable benchmarking and differentiated comparisons of methods that use a standardized tuning regime or a custom regime within one framework.
\RelArena enables developers to investigate isolated methodological improvements, custom and novel tuning regimes, or co-dependent improvements.
To support these different types of research, following TabArena, we categorize method submissions as ``models'' or ``systems'':
\begin{itemize}
    \item \textbf{Model submissions} follow a standardized tuning regime, allowing us to make claims about isolated methodological effects.
    In \RelArena, the standardized tuning regime includes random/grid search and early stopping on the validation set.
    A model submission only needs to declare a search space; \RelArena controls configuration sampling, run scheduling, and selection of the final candidate.
    Search spaces may not be dataset-specific beyond coarse differentiations based on dataset size, and we provide them for all implemented methods, mirroring the authors' choices where possible.
    \item \textbf{System submissions} may use custom tuning regimes, such as Bayesian optimization or conditional search steps.
    Comparing system submissions with each other allows us to understand which end-to-end pipeline performs best under the same input/output and time constraints.
    Between system and model submissions, we can compare final predictive performance, but efficiency and methodological improvements are not directly comparable because they may result from differences in the tuning regime.
\end{itemize}
This split between model and system submissions allows us to accommodate meaningful, novel research with system submissions while maintaining the ability to draw reliable research conclusions from model submissions.
Our formalization of model submissions constitutes a first, significant step towards comparable tuning in the \rellearning community.
Yet, fully aligning tuning regimes across methods and integrating all best practices remains an open research problem, mainly due to large differences in methodology, run times across methods and datasets, and preprocessing.
We discuss this in detail in Appendix~\ref{app:tuning}.
The long-term goal for benchmarking \rellearning should be to converge on best practices for tuning that make model development research meaningful for practitioners, while using system submissions to track the best end-to-end pipelines.

\paragraph{Unified evaluation regime and data state.} Through \RelArena's unified model API, we ensure that all methods receive the same data state during training, tuning, and evaluation, preventing any possible drift between evaluation regimes (\eg, as described in \cref{sec:problems:not_comparable}).

\paragraph{Strong set of baselines.}
We chose our initial \RelArena baselines based on their authors' self-reported performance on RelBench.
\RelArena implements GNN-based methods, such as GraphSAGE~\cite{rdl_fey}, RelGT~\cite{relgt} and RelGNN~\cite{relgnn}; relational foundation models, such as RT-PluRel~\cite{rt,plurel}; tabular aggregation-based approaches, such as RDBLearn~\cite{rdblearn} and \TabPFNRel~\cite{tabpfn_3_model_report}; and pragmatic baselines, such as the learning-free constant predictors, which output each task's optimal global or per-entity constant (\cref{sec:results}).
Of these, RT-PluRel is a system submission, and the rest are model submissions.
RT-PluRel uses the relational transformer model~\cite{rt,rt-j} pretrained on PluRel-generated~\cite{plurel} synthetic data and fine-tuned on the given task with a custom, sequential tuning regime (see Appendix~\ref{app:rt-plurel} for details).
\\
The current set of baselines does not reflect the complete state of \rellearning methods. We discuss this further in \Cref{sec:open_issues}. 
Yet, to the best of our knowledge, \RelArena presents the most comprehensive unified evaluation of \rellearning baselines to date. 
Together with the community, we will continue to grow the set of baselines. 

\paragraph{Shared evaluation tooling.} We integrate \RelArena directly with TabArena's \texttt{bencheval} package~\cite{tabarena}, the evaluation library behind the TabArena leaderboard. Given a results table produced by \RelArena, \texttt{bencheval} lets us compute a range of aggregate statistics and plots, including average ranks, critical-difference diagrams, leaderboards with bootstrapped Elo ratings (including confidence intervals), pairwise win-rate matrices, and normalized and baseline-relative scores. 
\\
As relational benchmarking matures and scales, the importance of trustworthy aggregate statistics will only grow. 
Likewise, sharing evaluation methodologies across communities is crucial to advancing evaluation research and unifying communication with practitioners. 

\subsection{\TabPFNRel}
\label{sec:contribution:tabpfnrel}

\begin{figure}[t]
   \centering
   \includegraphics[width=\linewidth]{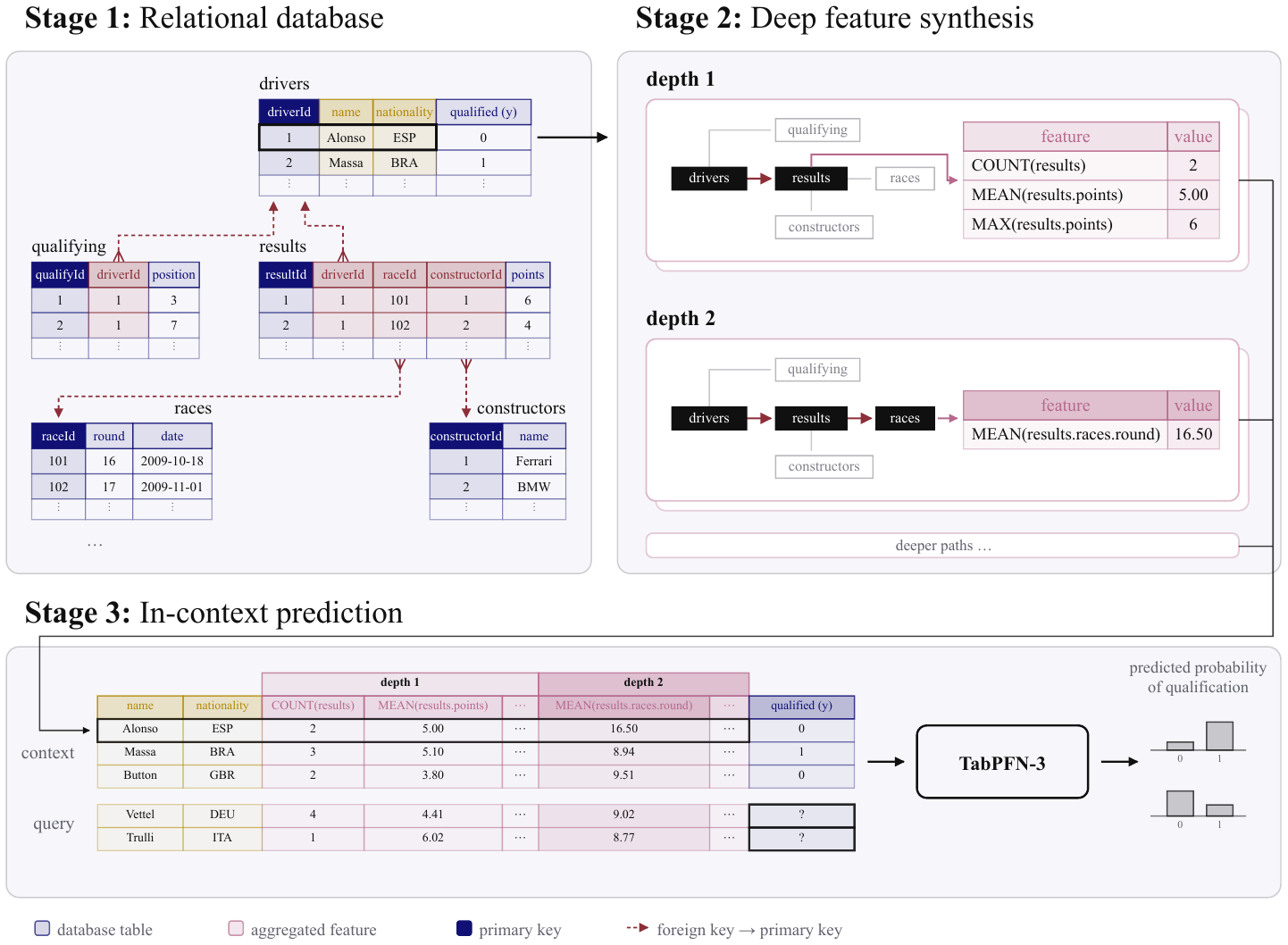}
\caption{The \TabPFNRel harness, illustrated on
\texttt{rel-f1/driver-top3}. Building on RDBLearn~\cite{rdblearn, rdblearncompanion}, \TabPFNRel uses deep feature
synthesis~\cite{featuretools} to aggregate relational features along
foreign-key paths into a flat table. \TabPFNThree then predicts query labels
in-context from labelled context rows. \Cref{sec:contribution:tabpfnrel}
details our improvements over the RDBLearn pipeline.}
   \label{fig:pipeline}
\end{figure}

\TabPFNRel is our relational harness for \TabPFNThree, our most recent \ac{tfm}. We visualize the harness in Figure \ref{fig:pipeline}. It builds upon RDBLearn~\cite{rdblearn, rdblearncompanion}, which in turn builds upon the popular \texttt{featuretools}~\cite{featuretools, featuretoolsrepo} library. As demonstrated in \cref{sec:results}, \TabPFNRel is currently the top-ranked model on \RelArena, demonstrating again~\cite{rdblearn, rdblearncompanion}
that flattening-based approaches to \rellearning can be strong baselines. \TabPFNRel inherits RDBLearn's core recipe: Each relational prediction task is converted into a flat table by exhaustively aggregating along all join paths implied by the schema's primary--foreign key relationships up to a maximum depth $d$ (deep feature synthesis~\cite{featuretools}), with $d$ tuned per task ($d \in \{2,3,4\}$). \TabPFNRel improves upon RDBLearn by:

\begin{itemize}
    \item \textbf{Improved tuning regime:} We resolve data drift issues that previously occurred during tuning by ensuring that the database used during tuning (the inner split) is frozen at the validation cut-off, mirroring how the final evaluation (the outer split) freezes it at the test cut-off, as outlined in \cref{sec:problems:not_comparable}. We note that as tuning is automated by \RelArena, all baseline methods, including RDBLearn, now benefit from this improvement.

    \item \textbf{Improved \acs{tfm} backbone:} \TabPFNThree is both more capable and scalable than previous \acp{tfm}~\cite{tabpfn_3_model_report} and was co-developed with the relational harness in mind. We make full use of \TabPFNThree's improvements by increasing the number of rows fed into the \ac{tfm} by an order of magnitude and replacing the set of \ac{tfm} backbones with \TabPFNThree. Even with the larger context size, \TabPFNRel's runtime stays comparable, both due to the removal of one tuning axis (RDBLearn additionally selects among three \ac{tfm} backbones in its sweep) and a more scalable \ac{tfm} architecture.
    \item \textbf{Support of text features:} We extend the RDBLearn recipe by re-attaching text columns from the entity table after the \texttt{featuretools} preprocessing. We then make use of \TabPFNThreePlus' superior ability to handle text features~\cite{tabpfn_3_model_report} to further boost performance. We note that this feature is only available via the TabPFN API. 
    For users who are not able to use the API, we also release a text-free version of \TabPFNRel, which users can run without querying the API.

    \item \textbf{Better context selection:} Lastly, we introduce a novel context selection regime, which replaces the random subsampling used in RDBLearn. By trading off recency and diversity among estimators' contexts, we are able to improve the harness's performance at no additional runtime cost. Additionally, after determining the optimal featurization depth $d$ on the validation set, we also re-use validation examples as additional context for making predictions on the test set. Due to the temporal nature of the forecasting tasks, the more recent validation examples are often particularly informative for making predictions.
\end{itemize}

\subsection{Relational Predictive Interface: \RPI}

Complementing \RelArena and \TabPFNRel, we release an initial version of our Relational Predictive Interface, \RPI, to enable the application of \rellearning methods, including \TabPFNRel, to real-world tasks.
\RPI is fully open-source, model-agnostic, and directly integrated into \RelArena, allowing users to run any \RelArena baseline, including hyperparameter tuning, on their own relational prediction problem in two lines of code. It generalizes the process used to generate the entity-level forecasting tasks of RelBench v1~\cite{relbenchv1}, but replaces custom task-generation code with a declarative interface: the database and the prediction task are specified entirely in YAML configuration files, without writing Python, turning a collection of CSV or Parquet files into a \RelArena task. To make \RPI as accessible as possible, we provide extensive documentation, an agent skill, example specifications for all 21 entity-level tasks in RelBench v1, and a Kaggle-based example.

We acknowledge that designing an interface for specifying relational prediction problems remains an open research question, in part because the requirements of practitioners are not yet fully understood. For this initial version of \RPI, we therefore adopt a deliberately expressive design aimed at researchers and early-adopting practitioners. The interface currently includes only limited safeguards against task mis-specification. This design reflects a broader tension between expressivity and usability: we expect that no single interface will be optimal for all settings and that different Pareto-optimal designs may serve different users, tasks, and levels of expertise. At the same time, the expressivity of \RPI is intentionally constrained to entity-level forecasting tasks in its current version to ensure compatibility with \RelArena. As a result, not every predictive task over a relational database can currently be represented within the framework. We are committed to working with the community to better understand the underlying trade-offs and to improve the accessibility, reliability, and expressivity of future versions of \RPI. In particular, we have recently become aware of similar, independent efforts to establish an interface for specifying entity-level forecasting tasks and are planning to unify \RPI's specification standard with this effort in the future.

\section{Results}
\label{sec:results}

\subsection{Experimental setup}

\paragraph{General setup.} We report the state of \RelArena at release over the
21 entity-level RelBench v1 tasks at a single seed, once for model submissions (under the tuning regime of
Appendix~\ref{app:tuning}) and once including system submissions in addition. We evaluate six model submissions: \TabPFNRel (API), which runs the hosted model with text features, and \TabPFNRel (OSS), which runs the open-source model without them; RDBLearn~\cite{rdblearn} (v1)\footnote{Our RDBLearn evaluation excludes LimiX, one of the three tabular foundation models considered in the original RDBLearn sweep, because it is not available as an installable PyPI package and would require vendoring an additional codebase. This may modestly disadvantage RDBLearn relative to its full configuration while also reducing its tuning cost. RDBLearn v1.1~\cite{paramfreeviable} was released after our baseline cut-off and will be evaluated in a future \RelArena release.}, GraphSAGE~\cite{rdl_fey, relbenchv1} (also known as the ``RDL'' baseline), RelGT~\cite{relgt}, and RelGNN~\cite{relgnn}.
\\
Moreover, for completeness, we include a common trivial LightGBM baseline, which we included in the rank and Elo computation but omitted from the figures, with per-task scores in \Cref{tab:raw_auroc,tab:raw_mae}. This trivial LightGBM baseline operates only on entity-level features~\cite{relbenchv1}, using the entity's own columns and the anchor timestamp but no aggregation over related tables. Likewise, we include two learning-free predictors, which predict the optimal constant for the task's primary metric (the positive-class rate for classification, the median for MAE), either globally (\texttt{constant-global}) or per entity (\texttt{constant-per-entity}).
\\
We further evaluate RT-PluRel~\cite{rt, plurel} with a custom tuning regime as a system submission. The distinction between submission types is explained in \cref{sec:contribution:relarena}.
We fit and evaluate every method through the \RelArena framework.

\paragraph{Known limitations.} We acknowledge that standardizing tuning across methods is not fully solved in \RelArena. In particular, while we made efforts to align runtimes across the different \RelArena baselines, significant variance remains, which could advantage more expensive methods.
We discuss this in detail in Appendix~\ref{app:tuning}, but note that \TabPFNRel does not have a significantly higher runtime than the other model submissions.

\paragraph{Possible conflicts of interest.} Releasing a benchmarking framework in addition to a model evaluated on said framework constitutes a conflict of interest. As stated in \Cref{sec:contribution:relarena}, \RelArena currently uses the most widely used set of \rellearning tasks~\cite{relbenchv1}, while applying benchmarking standards from the most widely used tabular benchmarking framework~\cite{tabarena}. We explain and motivate the methodological choices underlying \RelArena in \cref{sec:problems} and \cref{sec:contribution:relarena}; these choices try to reflect established practices rather than favoring any particular method. Moreover, by open-sourcing \RelArena, \TabPFNRel, and the custom RT-PluRel protocol, we make the benchmark design and our method implementations fully available for independent scrutiny and reproduction. This transparency allows questionable or potentially biased design choices, should any exist, to be readily identified, challenged, and evaluated under alternative benchmark configurations.

\subsection{Elo rankings on \RelArena}
\label{sec:results:eloranking}

\begin{figure}[h]
  \centering
  \includegraphics[width=\linewidth]{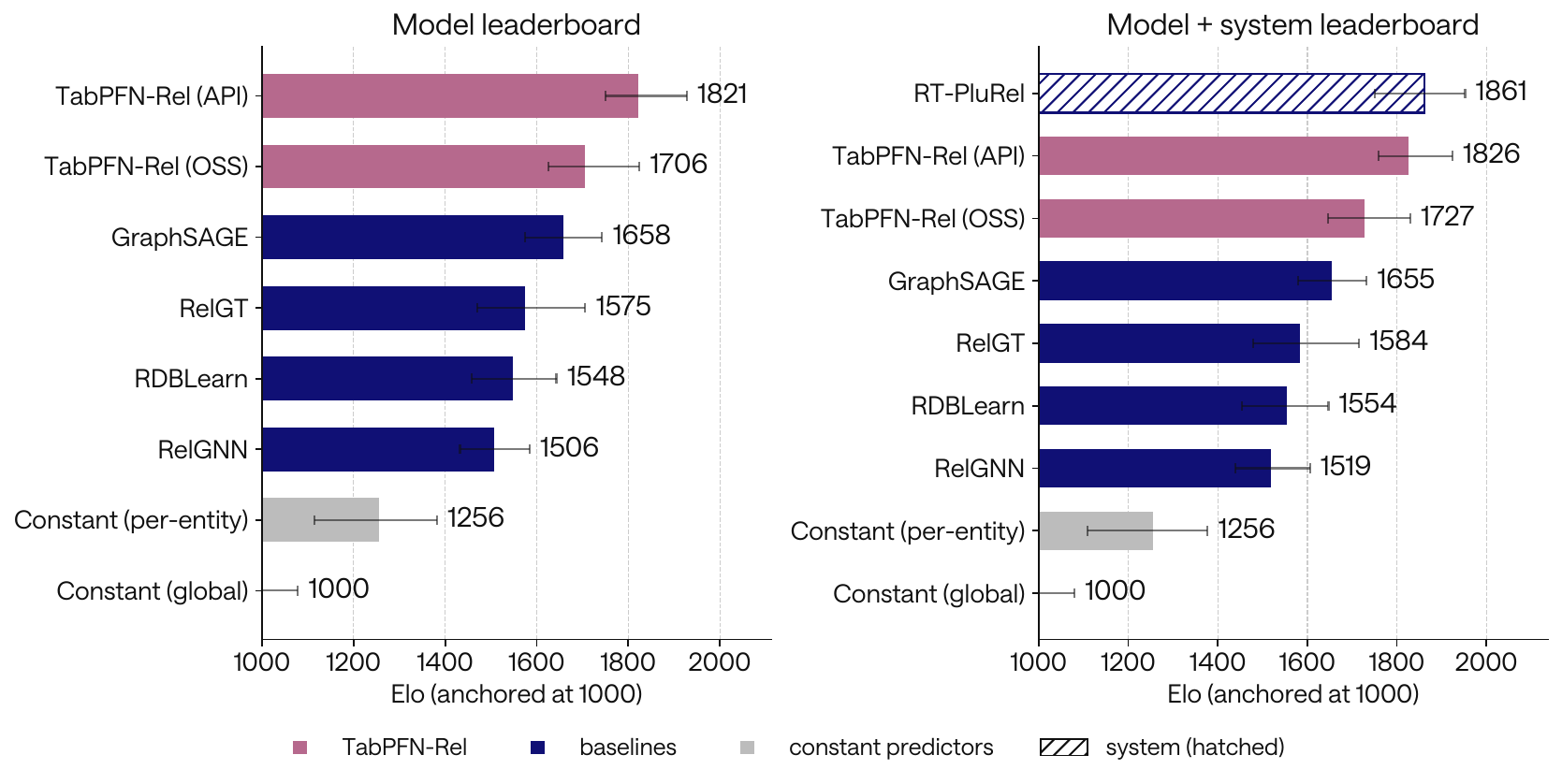}
  \caption{\textbf{Elo over the 21 \RelArena tasks: the model leaderboard (left) and the combined model + system leaderboard (right).} Entries marked as systems (hatched) comply with \RelArena's data states and evaluation regime but not with its standardized tuning regime (Appendix~\ref{app:tuning}).
  Each panel reports slightly different Elo for the same method, since Elo ratings are relative.
  \TabPFNRel ranks first among models with the standardized tuning regime; the system submission RT-PluRel ranks first in end-to-end performance.}
  \label{fig:elo}
\end{figure}

\Cref{fig:elo} presents the Elo ratings over \RelArena, computed using TabArena's \texttt{bencheval} \cite{tabarena}. Pairwise task outcomes across all method pairs are fit via maximum likelihood under the Bradley-Terry model~\cite{bradleyterry1952, chiang2024chatbot, tabarena}. Ratings are anchored to the global constant predictor at $1000$ points (where a $400$-point gap implies a $\approx 91\%$ win probability), with confidence intervals derived from bootstrapped battles. Appendix~\ref{app:raw} elaborates on the full results, presenting also dataset-level raw scores.

\paragraph{Main Results.} \Cref{fig:elo} presents the main leaderboards of \RelArena, separating model submissions from system submissions.
Model submissions are tuned under \RelArena's standardized, compute-matched tuning regime (Appendix~\ref{app:tuning}). System submissions respect \RelArena's data states and evaluation regime but bring their own tuning protocol. RT-PluRel~\cite{rt, plurel} is the first such system, using a custom, sequential tuning regime. 
Within models with the same tuning regime, \TabPFNRel ranks first. 
When including systems in the comparison, RT-PluRel with its custom tuning achieves the highest end-to-end predictive performance. 
RT-PluRel's strong performance shows the value of researching co-dependent model and tuning-regime improvements, while also prompting further investigation into how to transfer this insight to other models.

\paragraph{Tabular models are actually highly competitive.}
Contrary to prevailing (theoretical) beliefs in the \rellearning community, tabular models such as the \TabPFNRel variants and RDBLearn are competitive with relational deep learning baselines.
This adds to the growing evidence~\cite{rdblearn, rdblearncompanion, relagent, paramfreeviable} that flattening the database into a table is a strong strategy that can compete with relational deep learning models.
At the same time, relational models offer unique modeling advantages, raising the exciting question of how to combine the best of both worlds.

\paragraph{All methods are expensive to run.} Table~\ref{tab:runtime_stats} in Appendix~\ref{app:tuning} presents the runtimes of \RelArena. The single-seed leaderboard in \Cref{fig:elo} required hundreds of hours of wall-clock time. For more expensive datasets, some methods are prohibitively slow, making them impractical for many real-world applications. Our method is not exempt: \TabPFNRel requires a feature-synthesis pass over the database before it can be fit at all, and this CPU-bound preprocessing might dominate runtime depending on the dataset and hardware used. Other competitive methods are generally no cheaper to run: excluding preprocessing, which can be significantly more expensive for TabPFN-Rel, RT-PluRel's runtime is on average five times slower than TabPFN-Rel (API) and 30 times slower than TabPFN-Rel (OSS).
We believe that reducing this cost is key for this field to mature to industry adoption.

\paragraph{Text columns are critical for some tasks.}
Running \TabPFNRel without text features, as the OSS variant does, moves it from an Elo of 1821 to 1706, a drop larger than the gap between GraphSAGE and RelGT. 
The effect is not spread across the benchmark but dominated by two tasks: \texttt{rel-event/user-ignore} and \texttt{rel-avito/user-clicks}. 
Text is decisive where a task carries informative free-text columns and close to irrelevant elsewhere, as seen in Multimodal Tabular Learning benchmarks \cite{blayer2026strable, arazi2026multabench, arazi2026tabstar, purucker2026beyond}.

\paragraph{Constant predictors are not trivially beaten.}
\texttt{constant-per-entity} predicts each entity's own optimal constant from its training history, using no features and no model, and is nevertheless better than RelGNN and RelGT on 4 tasks each. \TabPFNRel and RT-PluRel are the only methods that exceed it on all 21. In addition, on the \texttt{rel-stack/post-votes} task, almost no method is significantly better than the constant predictor.

\subsection{Tunability in \rellearning}
\label{sec:results:tuning}

\begin{figure}[t]
  \centering
  \includegraphics[width=0.75\linewidth]{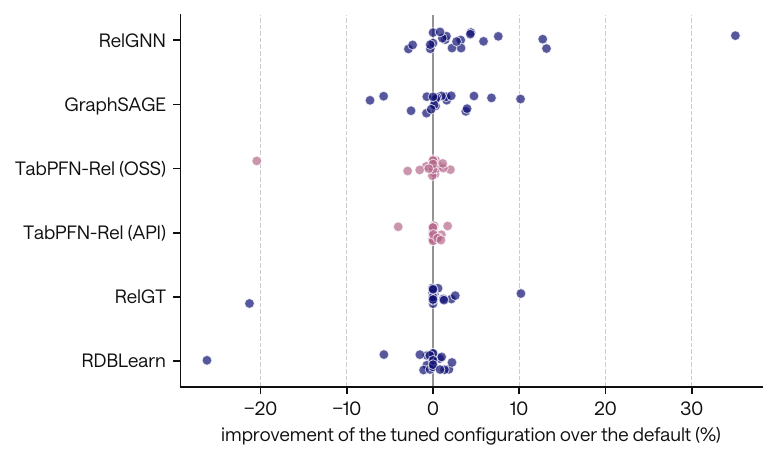}
  \caption{\textbf{Improvement of each method's tuned configuration over its own
  default.} One point per task and method, signed so that positive means tuning helped.}
  \label{fig:tuning}
\end{figure}

\Cref{fig:tuning} shows the improvement of tuning across all method–task pairs. We exclude RT-PluRel because it does not track the information required for this isolated investigation, as it is a system submission.
We observe that RelGNN benefits significantly from tuning, while GraphSAGE shows smaller improvements. By contrast, for most methods in \RelArena our tuning does not appear to lead to reliable improvements. \Cref{tab:tuning_counts} in the Appendix presents the breakdown per method. 
\\
The results have interesting implications for \TabPFNRel and other flattening-based approaches. \TabPFNRel searches only over aggregation depth (Appendix~\ref{app:tuning}), with the default corresponding to the shallowest depth considered. This default remains competitive with the deeper configurations selected by validation, suggesting that much of the predictive signal the method exploits is already accessible at depth two. Increasing the aggregation depth increases the number of join paths that must be materialized exponentially, leading to equivalently higher computational costs on larger databases. These results therefore suggest that simplifying the featurization procedure may be a promising avenue for improving the efficiency of \TabPFNRel without substantially compromising predictive performance. A more systematic investigation of novel tuning strategies and their interaction with computational efficiency in \rellearning remains open for future work. 

\section{Open Issues}
\label{sec:open_issues}

While our contributions address several of the challenges identified in~\cref{sec:problems}, important open issues remain. Some, such as expanding the set of baselines included in \RelArena, can be addressed through continued framework development. Others raise broader research questions concerning the design, evaluation, and practical applicability of relational learning methods. In this section, we outline these challenges both to clarify the current limitations of \RelArena and to identify promising directions for future work. Addressing them will require collaboration with the broader research community, and we hope that \RelArena can provide a foundation for such efforts.

\paragraph{Running models is difficult.} Running \rellearning methods remains far more expensive and error-prone than running tabular ones. Most competitive methods require hours of CPU-bound preprocessing per dataset before model training or inference. Examples include deep feature synthesis for flattening-based methods like RDBLearn and \TabPFNRel, or graph materialization and tokenization for GNN-based methods like RelGT and RelGNN. Including preprocessing in model training is possible but inefficient and impractical because CPU-based preprocessing and GPU-based model training have different hardware requirements. Additionally, runtime can vary by several orders of magnitude between methods. Both factors make fully standardizing tuning regimes across methods an open problem, which we discuss in Appendix~\ref{app:tuning}. Beyond benchmarking, broad adoption of \rellearning methods requires making methods significantly cheaper and easier to run.

\paragraph{Data quality requires further investigation.} With the release of \RelArena, we have mainly focused on standardizing benchmarking on a pre-existing set of databases and tasks, namely the entity-level RelBench v1 tasks. While the selected tasks remain the most popular in the community, we believe it will be important to understand whether the underlying databases and tasks are of sufficient quality and representative of the tasks practitioners need to solve in their day-to-day work.

\paragraph{Missing baselines.} As highlighted in \cref{sec:contribution:relarena}, it is likely that competitive baselines are missing from \RelArena for many possible reasons, such as differences in evaluation regimes, methods being released after our baseline cut-off date, or a lack of awareness on our part. The \RelArena team will continuously work with authors to make the set of included baselines as representative as possible. In particular, we make it easy for authors to add their methods to \RelArena by providing extensive documentation on the process as well as skill files for agentic coding tools. Stronger tabular baselines are likewise missing: we do not yet report baselines that lift traditional tabular learners to \rellearning through automated feature engineering, as discussed in \cref{sec:problems}. Establishing how strong such baselines can be, and how much of the gap to specialized \rellearning methods they close, is future work.

\paragraph{Disparate task types.} \RelArena currently limits itself to entity-level forecasting tasks~\cite{relbenchv1}. While these tasks remain the most popular, the community has introduced many task types, including recommendation tasks~\cite{relbenchv1}, entity attribute prediction~\cite{4dbinfer}, relationship attribute prediction~\cite{4dbinfer}, foreign key prediction~\cite{4dbinfer}, and autocomplete tasks~\cite{relbenchv2}. While many of these task types seem similar, they often have subtle differences that make them mutually incompatible, making cross-task method transfer difficult. We plan to extend \RelArena to additional task types once we have a sufficient understanding of which task types represent real-world predictive tasks practitioners need to solve. Initial investigations suggest that not all task types fulfill this constraint equally. 

\paragraph{Timestamp boundaries are not standardized.} Following prior work, GNN-based baselines as well as the RT-PluRel system implemented in \RelArena use rows at the entity's test timestamp as additional context, if test cut-off and entity timestamp align, while flattening-based approaches like TabPFN-Rel and RDBLearn do not. While this is currently permissible under \RelArena's API, this might disadvantage the latter set of methods. At this time, it is unclear to the authors whether this additional context is permissible in general, only for certain tasks, or not at all, or how much it influences model predictions. We plan to align these boundaries between methods in future releases.

\section{Outlook}

The release of \RelArena, \TabPFNRel, and \RPI marks only the first milestone of Prior Labs' long-term commitment to advancing the state of \rellearning. 
For \RelArena, we aim to work with the academic and open-source community to add baselines, further standardize tuning regimes, and address open questions about task types and data quality. The \TabPFNRel harness will be further co-developed with our upcoming model releases, as well as based on community feedback. Additionally, we are exploring further ideas to apply Prior Labs' expertise in building foundation models for structured data to the relational domain. 
Lastly, we will iterate \RPI based on community feedback to make \rellearning approaches as accessible as possible to practitioners, closing the gap between academia and industry.

\textbf{To conclude: towards non-dogmatic \rellearning.}
With \RelArena, \TabPFNRel, and \RPI, we are advancing open and reproducible \rellearning. 
We are optimistic that our contributions can be an inflection point for the community, accelerating research progress and real-world impact. 
And, we strongly believe that this can only become reality when we, as the \rellearning community, avoid dogmatic practices. 
A non-dogmatic \rellearning community may be the key to lasting success by enabling more welcoming, rigorous, and collaborative research.  
Different sub-communities, such as graph, relational, or tabular researchers, would strongly benefit from working together and evaluating on the same, fair benchmarks.

\clearpage

\bibliographystyle{unsrtnat}
\bibliography{references_frozen,references_manual}

\clearpage
\appendix

\section{Authorship and Contributions}
\label{app:authorship}

Adrian Hayler is listed first among the core contributors. The order of the remaining core contributors, Klemens Fl\"oge and Alan Arazi, was randomized. Rishabh Ranjan contributed the RT-PluRel system submission. Aside from this, the external contributors Rishabh Ranjan and Jure Leskovec, as well as our scientific advisors listed below, contributed scientific advice only. They did not participate in the design or implementation of \RelArena and contributed no intellectual property to it. Lennart Purucker, Frank Hutter, and Noah Hollmann led and supervised the project. The rest of the Prior Labs team is listed below, appearing in random order per group.

\textbf{Research/Model Dev:} Adrian Hayler, Klemens Fl\"oge, Alan Arazi, Lennart Purucker, Oscar Key, Philipp Jund, Vahid Balazadeh, Felix Birkel, Mihir Manium, Léo Grinsztajn, Arthur Cahu, Siyuan Guo, Tobias Schroeder, Jonas Kübler, David Salinas, Jan Hendrik Metzen, Anurag Garg, Jake Robertson, Benjamin Jäger, Nick Erickson, Simon Bing.

\textbf{Engineering/Platform:} Dominik Safaric, Simone Alessi, Brendan Roof, Georg Grab.

\textbf{Applied:} Philipp Singer, Eliott Kalfon.

\textbf{GTM:} Clara Cornu, Vitor Monteiro, Diana Kriuchkova, Tuana Çelik, Lilly Wehrhahn.

\textbf{Ops/People:} Kürşat Kaya, Jerry Chen, Lydia Sidhoum, Rylee Grace, Marie Salmon, Tomás Pereda, Kristina Collins.

\textbf{Scientific Advisors:}
Yann LeCun, 
Bernhard Schölkopf,
Madelon Hulsebos.

\textbf{Founders:} Sauraj Gambhir, Frank Hutter, Noah Hollmann.

\section{Tuning Regime in \RelArena}
\label{app:tuning}

This appendix expands on the tuning regime summarized in \cref{sec:contribution:relarena}. We describe how tuning is implemented in \RelArena, the runtime policy that governs it, and the remaining challenges in making tuning fully comparable across methods. Such comparability requires standardization along two dimensions:

\begin{enumerate}[label=(\roman*)]
\item \emph{How} methods are tuned: which data and metric are used for model selection, and which search spaces are admissible.
\item \emph{How much} methods are tuned: how much computation may be devoted to hyperparameter search.
\end{enumerate}

We believe that \RelArena largely addresses (i). By contrast, (ii) remains unresolved. Although the current runtime policy represents substantial progress, we believe that further iterations are needed for a fully satisfactory solution, likely requiring broader involvement of the academic community.

\paragraph{Tuning procedure.}
Each method registers a search space in a standardized format together with a default configuration. The search space is either sampled randomly, using the run seed, or specified as a small fixed grid whose configurations are evaluated in a predefined order. \RelArena then performs tuning automatically. For each configuration, it fits the method on the inner split's training data and evaluates it on the corresponding validation data using the task's primary metric. The configuration with the best validation score is selected and refit on the outer split for final evaluation. The default configuration is refit under the same protocol, so each method reports both an untuned and a tuned result.

Methods additionally specify whether the final fit combines the training and validation data or retains the validation split for early stopping, following the protocol used in the corresponding publication. Search spaces may not be tailored to individual datasets, except through coarse tiers based on dataset size. At present, only RelGT uses such tiers, allowing us to reproduce the authors' implementation. We are planning to re-evaluate whether any adjustment of search spaces, based on dataset size, is necessary.

\paragraph{Challenges in standardizing budgets.}
Two properties of current \rellearning methods make it difficult to standardize tuning compute. 

First, runtimes on individual datasets can be prohibitively long. For example, under the authors' nine-configuration tuning regime, RelGT requires more than 40 hours on an RTX 6000 Pro GPU for \texttt{rel-avito/user-visits}, even when excluding CPU-based preprocessing that itself takes several hours. By comparison, the tuned LightGBM baseline completes the same task in under 30 seconds.

Second, many competitive methods rely on CPU-bound preprocessing, which can often take hours per dataset. Examples include deep feature synthesis for flattening-based methods like RDBLearn and \TabPFNRel and graph materialization or tokenization for GNN-based methods like RelGNN and RelGT. We initially implemented all methods as end-to-end baselines whose preprocessing was executed directly within \RelArena. We ultimately had to abandon this design because preprocessing is predominantly CPU-bound, whereas most GPU nodes provide comparatively weak CPUs. At the scale of our experiments, using GPU nodes for hundreds of hours of CPU preprocessing would have been difficult to justify economically, both for industry practitioners and ourselves.

These observations suggest that most current \rellearning methods are not well suited to end-to-end execution on a single hardware configuration. \RelArena therefore permits methods to compute preprocessing artifacts once and cache them on disk before a run. The corresponding cache-generation scripts must be public, allowing others to reconstruct the caches and enabling reviewers to inspect them for errors such as label leakage. Cache generation is currently done primarily via free-form scripts that are not directly linked to the \RelArena framework. In particular, this means we currently do not track runtimes for cache generation, which also includes nuances such as how to amortize the runtime of preprocessing that is shared among the different tasks of one database.
We regard this as a limitation of the current release and plan to address it in future versions.

\paragraph{Current runtime policy.}
We currently impose a maximum total runtime of 24 hours per task, including preprocessing, measured on the largest tasks. Moderately larger search spaces are permitted on smaller tasks when their cost remains reasonable. In practice, all released baseline models, except RelGT, run for less than 12 hours per task. Because these constraints are still approximate, we ask method developers to exercise reasonable judgment, and we encourage open discussion. System submissions must conform to the same runtime constraints, but sit outside the standardized tuning regime: RT-PluRel is not tuned through \RelArena's trial budget at all --- it runs a single configuration and selects its training step and context internally during each fit. Its recorded runtimes reach roughly 16 hours per task, within the 24-hour constraint; they are listed separately in \cref{tab:runtime_stats}. We permit such submissions in the current early version of \RelArena, but may tighten the requirements on system submissions in the long term, depending on where the academic consensus settles.

\paragraph{Released configurations.}
\Cref{tab:runtime_stats} summarizes the trial budgets and runtimes of the released baselines. Trial budgets are specified per run rather than inferred automatically from the search spaces, and they reflect two different considerations. For methods with small fixed grids such as \texttt{rdblearn}, \texttt{tabpfn-rel} variants, or \texttt{relgt}, the budget is equal to the grid size, and the entire grid is evaluated with the default configuration first. For methods with sampled search spaces, the budget is chosen primarily according to computational cost. The inexpensive \texttt{lightgbm} baseline receives 30 trials, whereas the GPU-bound \texttt{graphsage} and \texttt{relgnn} baselines receive 4 and 10 trials, respectively. We note that the increased runtime of \texttt{tabpfn-rel-client} can be mostly attributed to calling TabPFN through the API rather than additional text processing. This task-independent overhead is reflected in the relatively moderate increase in maximum-over-median runtime.

RelGT is exceptional in two respects. First, it is the only RDL baseline whose tuning procedure is publicly available, with other methods either not disclosing their tuning regime or not using any tuning. We therefore tried to follow the authors' tuning regime as closely as possible in our initial implementation. Second, its grid depends on dataset size, which means that the effective number of trials on larger tasks may be smaller than the nominal budget. Once a more principled approach to standardizing tuning budgets is available, we intend to rerun all methods, including RelGT, under the revised regime.

\begin{table}[h]
\centering
\caption{Per-task runtime for methods on \RelArena at the time of release, aggregated over the 21 entity-level RelBench v1 tasks. $n_\text{max}$ indicates the maximum number of hyperparameter configurations a method tunes over per dataset. Runtimes cover all tuning trials plus both refits (default and selected configuration), but do not include the CPU-bound preprocessing, which is performed to varying degrees by all competitive methods, including \TabPFNRel and RDBLearn, which perform heavy CPU-bound feature engineering as part of the preprocessing. \textbf{The reported runtime numbers should therefore be seen as a lower bound on the actual runtime and are not directly representative of end-to-end method runtime.} The system submission \texttt{rt-plurel} is listed below the rule; it runs no tuning trials, so its runtime covers the fit and refit described in Appendix~\ref{app:rt-plurel}. We provide further details in Appendix~\ref{app:tuning}.}
\label{tab:runtime_stats}
\resizebox{\textwidth}{!}{%
\begin{tabular}{lrrrrrrr}
\toprule
Model & $n_\text{max}$ & Mean & Min & p25 & p50 & p75 & Max \\
\midrule
\texttt{constant-global} & 0 & 0.1\,s & 0.0\,s & -- & 0.0\,s & -- & 0.3\,s \\
\texttt{constant-per-entity} & 0 & 0.2\,s & 0.0\,s & -- & 0.1\,s & -- & 1.3\,s \\
\texttt{tabpfn-rel-client} & 3 & 76\,min & 18\,min & 72\,min & 88\,min & 94\,min & 108\,min \\
\texttt{tabpfn-rel-local} & 3 & 12\,min & 0.6\,min & 0.8\,min & 4\,min & 6\,min & 85\,min \\
\texttt{graphsage} & 4 & 47\,min & 2\,min & 9\,min & 35\,min & 83\,min & 141\,min \\
\texttt{rdblearn} & 6 & 11\,min & 0.5\,min & 2\,min & 7\,min & 11\,min & 50\,min \\
\texttt{relgt} & 9 & 511\,min & 40\,min & 175\,min & 273\,min & 466\,min & 2442\,min \\
\texttt{relgnn-es} & 10 & 73\,min & 1\,min & 6\,min & 28\,min & 99\,min & 336\,min \\
\texttt{lightgbm} & 30 & 14\,min & 0.1\,min & 0.2\,min & 2\,min & 28\,min & 53\,min \\
\midrule
\texttt{rt-plurel} & -- & 351\,min & 109\,min & 129\,min & 231\,min & 516\,min & 979\,min \\
\bottomrule
\end{tabular}}
\end{table}

\section{Comparison against self-reported results}
\label{app:comparison_to_self_reported}

\input{results_analysis/selfreported_elo}

In \cref{tab:elo_selfreported}, we compare results obtained through \RelArena with those reported by the original authors. We exclude RDBLearn from this analysis because results are not reported for 5 of the 21 RelBench v1 tasks.

Methods with weaker self-reported performance, such as
GraphSAGE, improve moderately under \RelArena, likely in part because they benefit from the standardized tuning procedure. By contrast, RelGNN's Elo decreases by nearly 450 points relative to its self-reported results. We believe this discrepancy is largely driven by differences in tuning: \RelArena's standardized tuning budget may be insufficient to explore RelGNN's huge configuration space as thoroughly as the authors' original tuning procedure (see \cref{sec:problems:not_reproducible} and Appendix~\ref{app:tuning}). In addition, because we had to re-implement RelGNN, implementation differences may contribute to the observed gap, despite following the authors' recommendations~\cite{relgnntrainingcodeissue}.

RelGT's Elo decreases more moderately, by around 200 points. The source of this discrepancy is less clear, as our implementation closely follows the authors' released implementation and already incurs substantially higher runtime than any other method in \RelArena (see Appendix~\ref{app:tuning}). \TabPFNRel, in contrast, improves slightly relative to the version reported in the \TabPFNThree model report~\cite{tabpfn_3_model_report}.

More generally, differences between self-reported and \RelArena results can arise for many reasons, including differences in evaluation and tuning regimes, implementation details, and errors in our own re-implementations. Going forward, we aim to work with the original authors to ensure that each method is represented as faithfully and fairly as possible within the constraints of \RelArena.

\section{Results tables per Task}
\label{app:raw}

\Cref{tab:raw_auroc,tab:raw_mae} report each method's test performance on all 21 \RelArena tasks, using AUROC for classification and MAE in each regression task's native units; every entry is the test score of the configuration selected on validation data. These per-task measurements determine the pairwise outcomes used to compute the Elo ratings and bootstrapped confidence intervals in \Cref{fig:elo}.

\input{results_analysis/tables}

\section{Effect of Tuning per Method}
\label{app:tuning_effect}

\Cref{tab:tuning_counts} counts, for each method, the tasks on which the
configuration selected on validation scores better, identically, or worse on
test than that method's own default. A task is unchanged almost exactly when
validation kept the default, so the middle column doubles as a count of the
tasks where the search found nothing worth taking.

\begin{table}[h]
\centering
\caption{\textbf{Effect of tuning, per method.} Number of the 21 tasks on which
the selected configuration beats, matches, or loses to that method's own default
on test. The constant predictors have no search space and so cannot move.}
\label{tab:tuning_counts}
\begin{tabular}{@{}lrrr@{}}
\toprule
Method & Better & Same & Worse \\
\midrule
RelGNN             & 15 &  2 & 4 \\
GraphSAGE          & 14 &  1 & 6 \\
LightGBM           & 14 &  1 & 6 \\
TabPFN-Rel (API)   &  9 &  7 & 5 \\
RDBLearn           &  8 &  6 & 7 \\
TabPFN-Rel (OSS)   &  8 &  5 & 8 \\
RelGT              &  7 & 11 & 3 \\
Constant (per-entity) &  0 & 21 & 0 \\
Constant (global)     &  0 & 21 & 0 \\
\bottomrule
\end{tabular}
\end{table}

\section{The RT-PluRel System Submission}
\label{app:rt-plurel}

RT-PluRel is the first system submission in \RelArena{}: it registers an
empty search space and runs a single configuration, performing all model
selection internally during each fit. It builds on the pre-trained relational
transformer~\cite{rt, plurel, rt-j}, an 85M-parameter model (12 blocks,
model dimension 512) that represents a relational database as a set of cell
tokens connected by column, row, and foreign-key attention, with text cells
embedded by a frozen sentence encoder (\texttt{all-MiniLM-L12-v2}). We use the
publicly released checkpoints pre-trained under the PluRel synthetic-data
protocol~\cite{plurel}, one per task type (classification and
regression).\footnote{\url{https://huggingface.co/stanford-star/rt-plurel}}
For each prediction, the model receives a context assembled from the entity's
relational neighborhood via breadth-first expansion and random-walk ranking;
the context is bounded by the split's cut-off timestamp, so tuning and
evaluation respect \RelArena{}'s data states (cf.\ \cref{sec:problems:not_comparable} on the timestamp boundary).

\paragraph{Sequential tuning regime.}
Each fit proceeds in two stages. \emph{Stage 1: fine-tuning with early
stopping.} The pre-trained checkpoint is delta-fine-tuned on the task's
training split---a zero-initialized additive delta on frozen pre-trained
weights, so that weight decay regularizes toward the pre-trained model---for up
to 50{,}000 steps (Muon optimizer, constant learning rate $5\cdot10^{-4}$,
batch size 256, stochastic weight averaging). Training batches mix context
configurations sampled from a grid over total context size
$\{128, 256, 512, 1024\}$ (in cells), local context size, neighborhood width
$\{16, 64, 256\}$, and a recency-preference flag. Every 100 steps, the averaged
weights are evaluated on 1{,}024 validation rows under two endpoint context
configurations; the best checkpoint under the task's primary metric is
retained, and training stops after 10{,}000 steps without improvement.
\emph{Stage 2: context selection.} With the selected checkpoint frozen, 60
context configurations from the same grid are scored on up to 4{,}096
validation rows with four context-sampling seeds each, and the best
configuration under the primary metric is selected.

\paragraph{Refit and prediction.}
Following \RelArena{}'s refit protocol, the model is then retrained from
the pre-trained checkpoint on the union of training and validation data,
scaling the selected step count by the ratio of dataset sizes, without any
further model selection. Test predictions use the selected context
configuration and ensemble eight context-sampling seeds, averaging raw outputs
before the final sigmoid or denormalization. Because both stages select on
validation data internally, RT-PluRel does not fit \RelArena{}'s
standardized trial-budget accounting and is therefore reported as a system
submission; see Appendix~\ref{app:tuning} for the runtime discussion.

\section{Data Provenance}
\label{app:provenance}

\RelArena distributes no data. All seven RelBench v1 databases are obtained at runtime through the \texttt{relbench} package, and each remains subject to the terms of its original source. \Cref{tab:provenance} lists the origin of each database, following the account of provenance given by RelBench~\cite{relbenchv1}. Anyone using \RelArena is bound by them directly and should consult the sources before relying on these databases for purposes of their own.

\begin{table}[h]
\centering
\footnotesize
\caption{Origin of the RelBench v1 databases evaluated in \RelArena.}
\label{tab:provenance}
\begin{tabular}{@{}l p{0.75\textwidth}@{}}
\toprule
Database & Origin \\
\midrule
\texttt{rel-amazon} & Amazon Review Data dump~\cite{amazonreviewdump, ni2019justifying} \\
\addlinespace
\texttt{rel-avito} & Kaggle Avito Context Ad Clicks competition~\cite{kaggleavito}\\
\addlinespace
\texttt{rel-event} & Kaggle Event Recommendation Engine Challenge~\cite{kaggleevent} \\
\addlinespace
\texttt{rel-f1} & Ergast Developer API, retrieved February 2024~\cite{ergastterms} \\
\addlinespace
\texttt{rel-hm} & Kaggle H\&M Personalized Fashion Recommendations competition~\cite{kagglehm} \\
\addlinespace
\texttt{rel-stack} & Stack Exchange data dump via the Internet Archive, November 2023~\cite{stackexchangedump} \\
\addlinespace
\texttt{rel-trial} & ClinicalTrials.gov, January 2024 snapshot~\cite{clinicaltrialsterms} \\
\bottomrule
\end{tabular}
\end{table}

\paragraph{Attribution for \texttt{rel-stack}.}
The \texttt{rel-stack} database is built from content contributed by Stack Exchange users under CC BY-SA 3.0. The license distributed with the dump sets out its attribution requirements for republishing that content, which \RelArena does not do. We nonetheless acknowledge the Stack Exchange communities and the individual contributors whose questions and answers make up this database, since their work is what the corresponding results measure.

\paragraph{Acknowledgment for \texttt{rel-event}.}
The Event Recommendation Engine Challenge data was made available for academic use with the explicit consent of its creators, and we join RelBench in thanking Allan Carroll for sharing it with the academic community~\cite{relbenchv1}. That consent was given to RelBench, and we use the database only as RelBench distributes it.

\paragraph{Retrieval route.}
RelBench states that its users obtain the \texttt{rel-avito}, \texttt{rel-hm} and \texttt{rel-event} data from Kaggle themselves, and identifies accepting the competition terms as part of that step~\cite{relbenchv1}.

\section{Acknowledgements}
\label{app:acknowledge}
We acknowledge the EuroHPC Joint Undertaking for awarding this project access to the EuroHPC supercomputers LUMI, hosted by CSC (Finland) and the LUMI consortium, and MareNostrum 5, hosted by the Barcelona Supercomputing Center (Spain), through EuroHPC AI Factory Large Scale Access calls (projects EHPC-AIF-2025LS17-014 and EHPC-AIF-2025LS14-031).

This work was supported by European Union's Horizon Europe research and innovation programme under grant agreement number 101214398 (ELLIOT).

\textbf{Disclaimer:} Funded by the European Union. Views and opinions expressed are however those of the author(s) only and do not necessarily reflect those of the European Union or the European Commission. Neither the European Union nor the European Commission can be held responsible for them.

% EU emblem rules: the flag must be at least 1 cm high and at least as large as
% any other logo shown with it; the flag fills 86% of the image height.
\begin{center}
    \includegraphics[height=1.2cm]{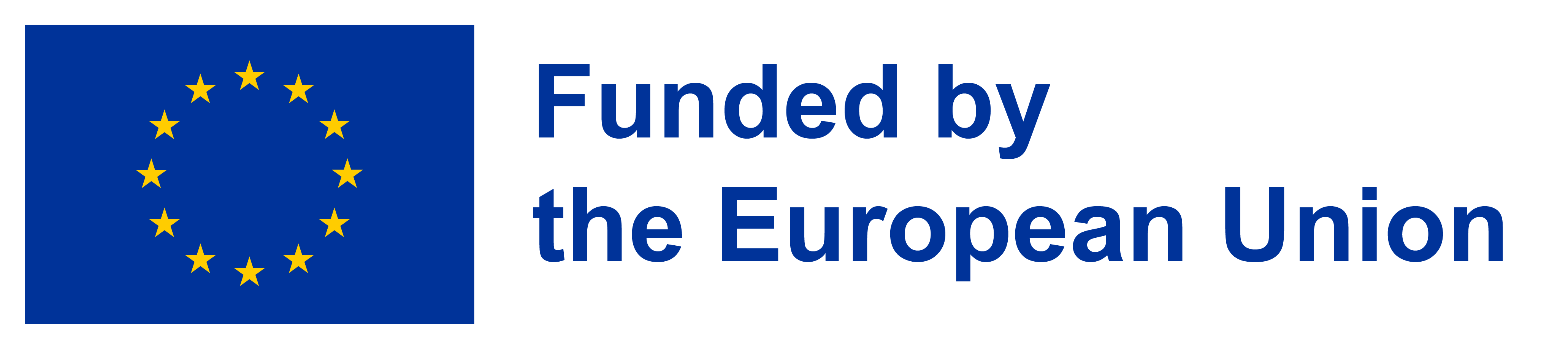}\hfill
    \includegraphics[height=1cm]{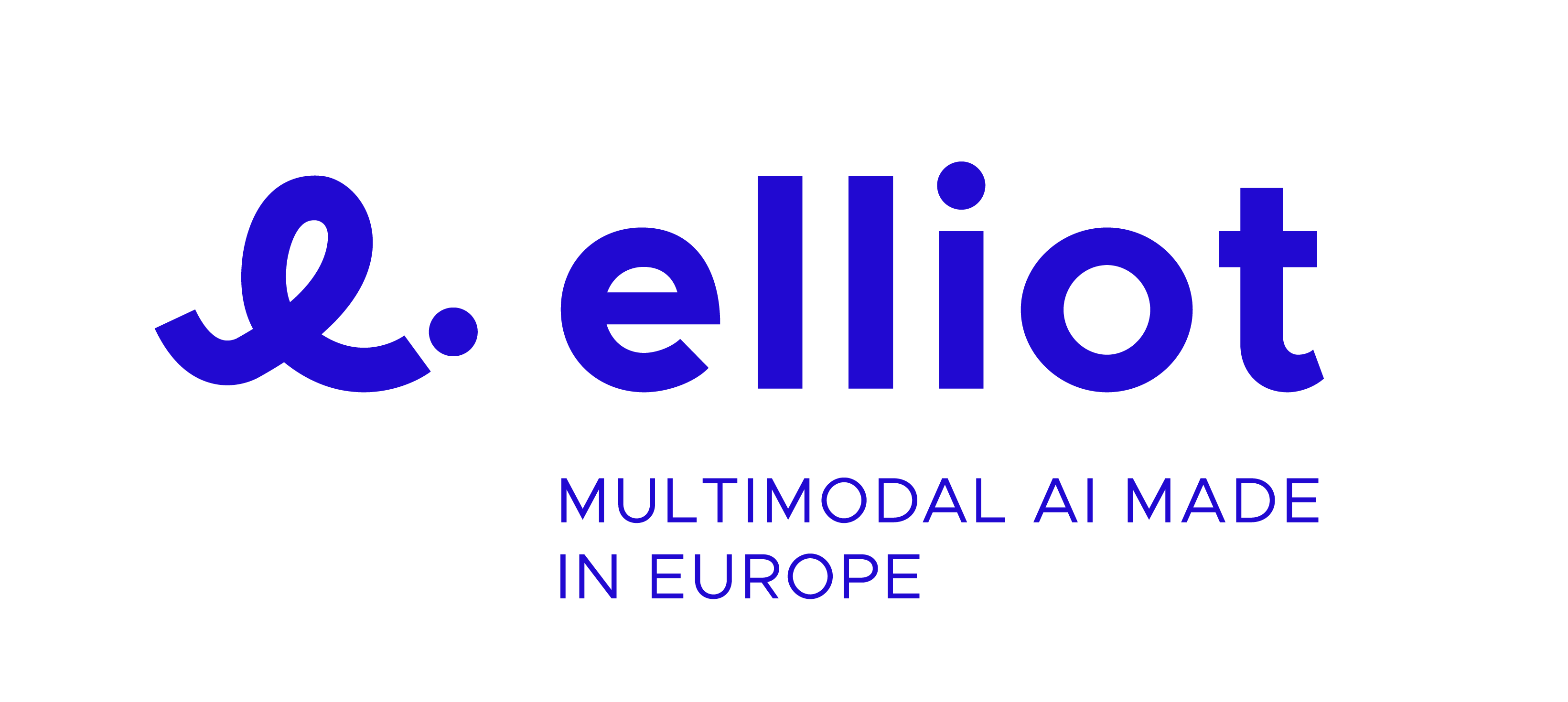}\hfill
    \includegraphics[height=1cm]{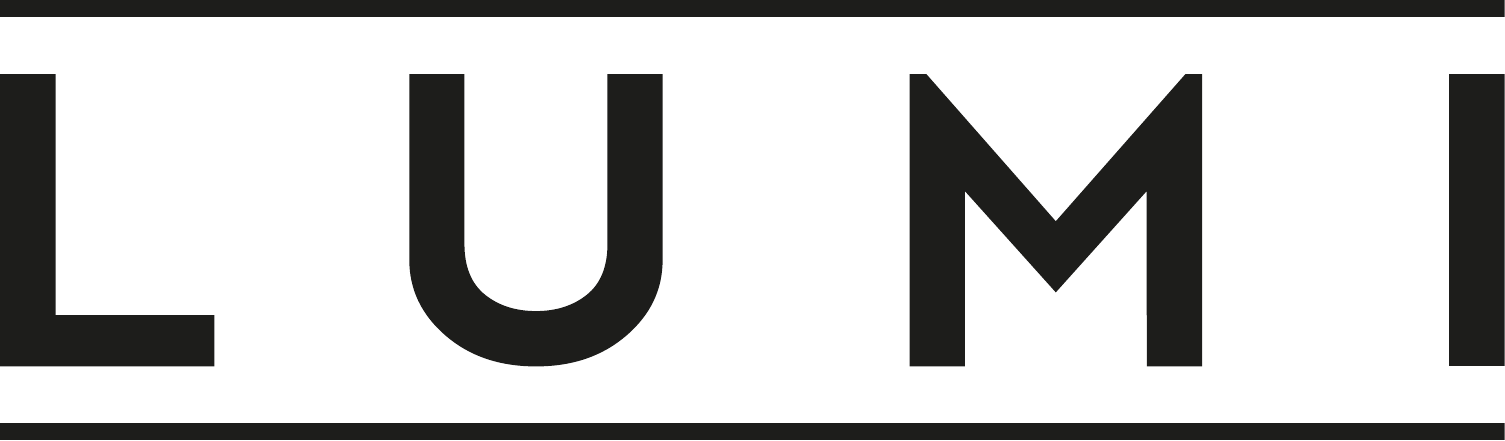}
\end{center}

\end{document}

%% file: macros.tex
\newcommand{\RelArena}{\texorpdfstring{RelArena-$\alpha$}{RelArena-α}\xspace}
\newcommand{\TabPFNRel}{TabPFN-Rel\xspace}
\newcommand{\RPI}{RPI\xspace}
\newcommand{\TabPFNThree}{TabPFN-3\xspace}
\newcommand{\TabPFNThreePlus}{TabPFN-3-Plus\xspace}
\newcommand{\rellearning}{relational learning\xspace}

\newcommand*{\eg}{e.g.\@\xspace}

\acrodef{rdb}[RDB]{relational database}
\acrodef{fe}[FE]{feature engineering}
\acrodef{tfm}[TFM]{tabular foundation model}

\newcommand{\cmark}{\textcolor{green!60!black}{\ding{51}}}
\newcommand{\xmark}{\textcolor{red!80!black}{\ding{55}}}

%% file: results_analysis/selfreported_elo.tex
\begin{table}[h]
\centering
\caption{\textbf{Comparison of Elo ratings from authors' self-reported results and results obtained under \RelArena's standardized evaluation framework.} Elo is computed on a joint board containing both sets of results, together with the constant baselines, and anchored at the global constant baseline. Because the two sets of results were obtained under different evaluation and tuning regimes, as discussed in \Cref{sec:problems}, these Elo differences should be interpreted as measures of discrepancy rather than differences in method quality. We exclude RDBLearn~\cite{rdblearn} because the authors do not report results for 5 of the 21 tasks, and RT-PluRel because no published results correspond to its \RelArena configuration: the RT and PluRel papers~\cite{rt, plurel} evaluate earlier versions of the code without per-task fine-tuning of the PluRel-pretrained model.}
\label{tab:elo_selfreported}
\begin{tabular}{lrrr}
\toprule
Method & Elo & CI$+$ & CI$-$ \\
\midrule
RelGNN (self-reported) & 1949 & $+114$ & $-76$ \\
\TabPFNRel{} (API) & 1851 & $+82$ & $-79$ \\
\TabPFNRel{} (v3 model report) & 1810 & $+116$ & $-76$ \\
RelGT (self-reported) & 1806 & $+94$ & $-82$ \\
GraphSAGE (\RelArena) & 1714 & $+90$ & $-94$ \\
GraphSAGE (self-reported) & 1691 & $+73$ & $-73$ \\
RelGT (\RelArena) & 1605 & $+127$ & $-113$ \\
RelGNN (\RelArena) & 1503 & $+79$ & $-72$ \\
LightGBM (\RelArena) & 1365 & $+104$ & $-131$ \\
LightGBM (self-reported) & 1324 & $+93$ & $-106$ \\
Constant (per-entity) & 1262 & $+124$ & $-139$ \\
Constant (global) & 1000 & $+88$ & $-185$ \\
\bottomrule
\end{tabular}
\end{table}

%% file: results_analysis/tables.tex
% Raw-score tables, generated by make_figures.py -- do not hand-edit.
% Each table is wrapped in \resizebox{\textwidth}{!}{...} -- needs graphicx.

\begin{table}[h]
\centering
\footnotesize
\setlength{\tabcolsep}{3.5pt}
\caption{Test AUROC per task for every method run in \RelArena (higher is better, best per task in bold). Columns, left to right: TPR = \texttt{tabpfn-rel-client}, TPR-OSS = \texttt{tabpfn-rel-local}, RT-P = \texttt{rt-plurel}, GSage = \texttt{graphsage}, RDBL = \texttt{rdblearn}, RGNN = \texttt{relgnn-es}, RGT = \texttt{relgt}, LGBM = \texttt{lightgbm}, Const-e = \texttt{constant-per-entity}, Const-g = \texttt{constant-global}.}
\label{tab:raw_auroc}
\resizebox{\textwidth}{!}{%
\begin{tabular}{lrrrrrrrrrr}
\toprule
Task & TPR & TPR-OSS & RT-P & GSage & RDBL & RGNN & RGT & LGBM & Const-e & Const-g \\
\midrule
\texttt{rel-amazon/item-churn} & 0.8280 & 0.8279 & \textbf{0.8327} & 0.8305 & 0.8195 & 0.7856 & 0.8238 & 0.6622 & 0.7289 & 0.5000 \\
\texttt{rel-amazon/user-churn} & 0.7086 & 0.7024 & \textbf{0.7135} & 0.7046 & 0.6844 & 0.6943 & 0.7019 & 0.5171 & 0.6342 & 0.5000 \\
\texttt{rel-avito/user-clicks} & 0.6752 & 0.6145 & 0.5834 & 0.6087 & \textbf{0.6788} & 0.6676 & 0.6444 & 0.5642 & 0.5041 & 0.5000 \\
\texttt{rel-avito/user-visits} & 0.6680 & 0.6688 & \textbf{0.6709} & 0.6658 & 0.6596 & 0.6487 & 0.6621 & 0.5293 & 0.6027 & 0.5000 \\
\texttt{rel-event/user-ignore} & \textbf{0.8787} & 0.7014 & 0.8476 & 0.7587 & 0.6644 & 0.8054 & 0.7815 & 0.7772 & 0.8399 & 0.5000 \\
\texttt{rel-event/user-repeat} & 0.7593 & 0.7693 & \textbf{0.7914} & 0.7846 & 0.7441 & 0.7546 & 0.7344 & 0.7483 & 0.7518 & 0.5000 \\
\texttt{rel-f1/driver-dnf} & \textbf{0.7322} & 0.7145 & 0.7315 & 0.7172 & 0.7146 & 0.7261 & 0.7117 & 0.7303 & 0.6993 & 0.5000 \\
\texttt{rel-f1/driver-top3} & 0.7714 & 0.7929 & 0.7589 & 0.7260 & 0.7801 & 0.7589 & \textbf{0.8108} & 0.7389 & 0.5565 & 0.5000 \\
\texttt{rel-hm/user-churn} & 0.7052 & \textbf{0.7057} & 0.7044 & 0.6985 & 0.6984 & 0.6820 & 0.6895 & 0.5901 & 0.6480 & 0.5000 \\
\texttt{rel-stack/user-badge} & 0.8804 & 0.8635 & \textbf{0.8916} & 0.8887 & 0.7711 & 0.6206 & 0.5743 & 0.5380 & 0.7890 & 0.5000 \\
\texttt{rel-stack/user-engagement} & 0.9060 & 0.9058 & 0.8968 & 0.9056 & 0.8587 & 0.9051 & \textbf{0.9067} & 0.8118 & 0.8267 & 0.5000 \\
\texttt{rel-trial/study-outcome} & \textbf{0.7647} & 0.7306 & 0.7235 & 0.6862 & 0.7212 & 0.6574 & 0.6685 & 0.7150 & 0.5000 & 0.5000 \\
\bottomrule
\end{tabular}}
\end{table}

\begin{table}[h]
\centering
\footnotesize
\setlength{\tabcolsep}{3.5pt}
\caption{Test MAE per task in the task's native units (lower is better, best per task in bold). Columns, left to right: TPR = \texttt{tabpfn-rel-client}, TPR-OSS = \texttt{tabpfn-rel-local}, RT-P = \texttt{rt-plurel}, GSage = \texttt{graphsage}, RDBL = \texttt{rdblearn}, RGNN = \texttt{relgnn-es}, RGT = \texttt{relgt}, LGBM = \texttt{lightgbm}, Const-e = \texttt{constant-per-entity}, Const-g = \texttt{constant-global}.}
\label{tab:raw_mae}
\resizebox{\textwidth}{!}{%
\begin{tabular}{lrrrrrrrrrr}
\toprule
Task & TPR & TPR-OSS & RT-P & GSage & RDBL & RGNN & RGT & LGBM & Const-e & Const-g \\
\midrule
\texttt{rel-amazon/item-ltv} & 46.8 & 47.8 & \textbf{43.0} & 49.2 & 49.0 & 52.5 & 48.7 & 55.8 & 65.4 & 64.2 \\
\texttt{rel-amazon/user-ltv} & 14.4 & 14.4 & \textbf{13.9} & 14.4 & 14.6 & 14.6 & 14.4 & 16.8 & 17.4 & 16.8 \\
\texttt{rel-avito/ad-ctr} & \textbf{0.0311} & 0.0314 & 0.0348 & 0.0390 & 0.0341 & 0.0426 & 0.0365 & 0.0412 & 0.0412 & 0.0431 \\
\texttt{rel-event/user-attendance} & 0.244 & \textbf{0.239} & 0.241 & 0.245 & 0.242 & 0.244 & 0.261 & 0.263 & 0.269 & 0.264 \\
\texttt{rel-f1/driver-position} & 3.769 & \textbf{3.762} & 3.818 & 4.011 & 3.889 & 4.266 & 4.766 & 4.106 & 4.104 & 4.399 \\
\texttt{rel-hm/item-sales} & 0.0605 & 0.0614 & \textbf{0.0403} & 0.0552 & 0.0671 & 0.0565 & 0.0532 & 0.0753 & 0.0780 & 0.0761 \\
\texttt{rel-stack/post-votes} & 0.0679 & 0.0680 & \textbf{0.0635} & 0.0649 & 0.0677 & 0.0679 & 0.0679 & 0.0661 & 0.0694 & 0.0679 \\
\texttt{rel-trial/site-success} & 0.413 & 0.386 & 0.410 & \textbf{0.325} & 0.486 & 0.340 & 0.370 & 0.438 & 0.441 & 0.462 \\
\texttt{rel-trial/study-adverse} & 39.8 & 42.6 & \textbf{32.7} & 44.3 & 44.0 & 46.3 & 44.1 & 44.6 & 57.5 & 57.5 \\
\bottomrule
\end{tabular}}
\end{table}

%% file: references_frozen.bib
@inproceedings{tabarena,
	title = {{TabArena}: {A} {Living} {Benchmark} for {Machine} {Learning} on {Tabular} {Data}},
	shorttitle = {{TabArena}},
	booktitle = {Advances in Neural Information Processing Systems ({NeurIPS}), Datasets and Benchmarks Track},
	url = {http://arxiv.org/abs/2506.16791},
	doi = {10.48550/arXiv.2506.16791},
	urldate = {2026-08-03},
	author = {Erickson, Nick and Purucker, Lennart and Tschalzev, Andrej and Holzmüller, David and Desai, Prateek Mutalik and Salinas, David and Hutter, Frank},
	year = {2025},
	note = {arXiv:2506.16791 [cs.LG]},
}

@misc{relagent,
	title = {{RelAgent}: {LLM} {Agents} as {Data} {Scientists} for {Relational} {Learning}},
	shorttitle = {{RelAgent}},
	url = {http://arxiv.org/abs/2605.07840},
	doi = {10.48550/arXiv.2605.07840},
	urldate = {2026-08-02},
	publisher = {arXiv},
	author = {Huang, Xingyue and Tichelman, Louis and Kim, Jinwoo and Olejniczak, Krzysztof and Ceylan, İsmail İlkan},
	month = may,
	year = {2026},
	note = {arXiv:2605.07840 [cs.LG]},
}

@inproceedings{lightgbm,
	title = {{LightGBM}: {A} {Highly} {Efficient} {Gradient} {Boosting} {Decision} {Tree}},
	volume = {30},
	shorttitle = {{LightGBM}},
	url = {https://proceedings.neurips.cc/paper_files/paper/2017/hash/6449f44a102fde848669bdd9eb6b76fa-Abstract.html},
	urldate = {2026-08-02},
	booktitle = {Advances in {Neural} {Information} {Processing} {Systems}},
	publisher = {Curran Associates, Inc.},
	author = {Ke, Guolin and Meng, Qi and Finley, Thomas and Wang, Taifeng and Chen, Wei and Ma, Weidong and Ye, Qiwei and Liu, Tie-Yan},
	year = {2017},
}

@inproceedings{rdl_fey,
	title = {Position: {Relational} {Deep} {Learning}: {Graph} {Representation} {Learning} on {Relational} {Databases}},
	shorttitle = {Relational {Deep} {Learning}},
	booktitle = {International Conference on Machine Learning ({ICML})},
	url = {http://arxiv.org/abs/2312.04615},
	doi = {10.48550/arXiv.2312.04615},
	urldate = {2026-08-02},
	author = {Fey, Matthias and Hu, Weihua and Huang, Kexin and Lenssen, Jan Eric and Ranjan, Rishabh and Robinson, Joshua and Ying, Rex and You, Jiaxuan and Leskovec, Jure},
	year = {2024},
	note = {arXiv:2312.04615 [cs.LG]},
}

@inproceedings{relgt,
	title = {Relational {Graph} {Transformer}},
	booktitle = {International Conference on Learning Representations ({ICLR})},
	url = {http://arxiv.org/abs/2505.10960},
	doi = {10.48550/arXiv.2505.10960},
	language = {en},
	urldate = {2026-06-10},
	author = {Dwivedi, Vijay Prakash and Jaladi, Sri and Shen, Yangyi and López, Federico and Kanatsoulis, Charilaos I. and Puri, Rishi and Fey, Matthias and Leskovec, Jure},
	year = {2026},
	note = {arXiv:2505.10960 [cs.LG]},
}

@inproceedings{rt,
	title = {Relational {Transformer}: {Toward} {Zero}-{Shot} {Foundation} {Models} for {Relational} {Data}},
	author = {Ranjan, Rishabh and Hudovernik, Valter and Znidar, Mark and Kanatsoulis, Charilaos and Upendra, Roshan and Mohammadi, Mahmoud and Meyer, Joe and Palczewski, Tom and Guestrin, Carlos and Leskovec, Jure},
	booktitle = {International Conference on Learning Representations ({ICLR})},
	url = {http://arxiv.org/abs/2510.06377},
	year = {2026},
}

@inproceedings{plurel,
	title = {{PluRel}: {Synthetic} {Data} unlocks {Scaling} {Laws} for {Relational} {Foundation} {Models}},
	author = {Kothapalli, Vignesh and Ranjan, Rishabh and Hudovernik, Valter and Dwivedi, Vijay Prakash and Hoffart, Johannes and Guestrin, Carlos and Leskovec, Jure},
	booktitle = {International Conference on Machine Learning ({ICML})},
	url = {http://arxiv.org/abs/2602.04029},
	year = {2026},
}

@inproceedings{rt-j,
	title = {Large-{Scale} {Pretraining} unlocks {Few}-{Shot} {Prediction} for {Relational} {Data}},
	author = {Ranjan, Rishabh and Kothapalli, Vignesh and Agarwal, Harshvardhan and Kanatsoulis, Charilaos I. and Upendra, Roshan Reddy and Palczewski, Tom and Guestrin, Carlos and Leskovec, Jure},
	booktitle = {2nd {ICML} Workshop on Foundation Models for Structured Data},
	url = {https://openreview.net/forum?id=oQINTd9din},
	year = {2026},
}

@misc{kumorfmv2,
	title = {{KumoRFM}-2: {Scaling} {Foundation} {Models} for {Relational} {Learning}},
	shorttitle = {{KumoRFM}-2},
	url = {http://arxiv.org/abs/2604.12596},
	doi = {10.48550/arXiv.2604.12596},
	urldate = {2026-07-31},
	publisher = {arXiv},
	author = {Hudovernik, Valter and López, Federico and Kocijan, Vid and Nitta, Akihiro and Lenssen, Jan Eric and Leskovec, Jure and Fey, Matthias},
	month = apr,
	year = {2026},
	note = {arXiv:2604.12596 [cs.LG]},
}

@inproceedings{relgnn,
	title = {{RelGNN}: {Composite} {Message} {Passing} for {Relational} {Deep} {Learning}},
	booktitle = {International Conference on Machine Learning ({ICML})},
	shorttitle = {{RelGNN}},
	url = {http://arxiv.org/abs/2502.06784},
	doi = {10.48550/arXiv.2502.06784},
	language = {en},
	urldate = {2026-06-16},
	author = {Chen, Tianlang and Kanatsoulis, Charilaos and Leskovec, Jure},
	year = {2025},
	note = {arXiv:2502.06784 [cs.LG]},
}

@misc{pql,
	title = {Predictive {Query} {Language}: {A} {Domain}-{Specific} {Language} for {Predictive} {Modeling} on {Relational} {Databases}},
	shorttitle = {Predictive {Query} {Language}},
	url = {http://arxiv.org/abs/2602.09572},
	doi = {10.48550/arXiv.2602.09572},
	urldate = {2026-06-23},
	publisher = {arXiv},
	author = {Kocijan, Vid and Sunil, Jinu and Lenssen, Jan Eric and Deb, Viman and Xe, Xinwei and Gomez, Federico Reyes and Fey, Matthias and Leskovec, Jure},
	month = feb,
	year = {2026},
	note = {arXiv:2602.09572 [cs.DB]. Presented at TaDA@VLDB 2026},
}

@misc{rdblearn,
	title = {{RDBLearn}: {Simple} {In}-{Context} {Prediction} {Over} {Relational} {Databases}},
	shorttitle = {{RDBLearn}},
	url = {http://arxiv.org/abs/2602.18495},
	doi = {10.48550/arXiv.2602.18495},
	urldate = {2026-07-31},
	publisher = {arXiv},
	author = {Zhang, Yanlin and Xu, Linjie and Gan, Quan and Wipf, David and Wang, Minjie},
	month = feb,
	year = {2026},
	note = {arXiv:2602.18495 [cs.DB]},
}

@inproceedings{4dbinfer,
	title = {{4DBInfer}: {A} {4D} {Benchmarking} {Toolbox} for {Graph}-{Centric} {Predictive} {Modeling} on {Relational} {DBs}},
	booktitle = {Advances in Neural Information Processing Systems ({NeurIPS}), Datasets and Benchmarks Track},
	shorttitle = {{4DBInfer}},
	url = {http://arxiv.org/abs/2404.18209},
	doi = {10.48550/arXiv.2404.18209},
	urldate = {2026-05-22},
	author = {Wang, Minjie and Gan, Quan and Wipf, David and Cai, Zhenkun and Li, Ning and Tang, Jianheng and Zhang, Yanlin and Zhang, Zizhao and Mao, Zunyao and Song, Yakun and Wang, Yanbo and Li, Jiahang and Zhang, Han and Yang, Guang and Qin, Xiao and Lei, Chuan and Zhang, Muhan and Zhang, Weinan and Faloutsos, Christos and Zhang, Zheng},
	year = {2024},
	note = {arXiv:2404.18209 [cs.LG]},
}

@inproceedings{redelex,
	title = {{REDELEX}: {A} {Framework} for {Relational} {Deep} {Learning} {Exploration}},
	booktitle = {Machine Learning and Knowledge Discovery in Databases ({ECML} {PKDD} 2025)},
	volume = {16014},
	shorttitle = {{REDELEX}},
	url = {http://arxiv.org/abs/2506.22199},
	doi = {10.1007/978-3-032-05981-9_26},
	language = {en},
	urldate = {2026-05-27},
	author = {Peleška, Jakub and Šír, Gustav},
	year = {2025},
	note = {arXiv:2506.22199 [cs.LG]},
	pages = {438--456},
}

@inproceedings{relbenchv2,
	title = {{RelBench} v2: {A} {Large}-{Scale} {Benchmark} and {Repository} for {Relational} {Data}},
	shorttitle = {{RelBench} v2},
	booktitle = {3rd Workshop on Navigating and Addressing Data Problems for Foundation Models ({DATA-FM}) at {ICLR} 2026},
	url = {http://arxiv.org/abs/2602.12606},
	doi = {10.48550/arXiv.2602.12606},
	urldate = {2026-05-22},
	author = {Gu, Justin and Ranjan, Rishabh and Kanatsoulis, Charilaos and Tang, Haiming and Jurkovic, Martin and Hudovernik, Valter and Znidar, Mark and Chaturvedi, Pranshu and Shroff, Parth and Li, Fengyu and Leskovec, Jure},
	year = {2026},
	note = {arXiv:2602.12606 [cs.LG]},
}

@inproceedings{relbenchv1,
	title = {{RelBench}: {A} {Benchmark} for {Deep} {Learning} on {Relational} {Databases}},
	booktitle = {Advances in Neural Information Processing Systems ({NeurIPS}), Datasets and Benchmarks Track},
	shorttitle = {{RelBench}},
	url = {http://arxiv.org/abs/2407.20060},
	doi = {10.48550/arXiv.2407.20060},
	urldate = {2026-05-22},
	author = {Robinson, Joshua and Ranjan, Rishabh and Hu, Weihua and Huang, Kexin and Han, Jiaqi and Dobles, Alejandro and Fey, Matthias and Lenssen, Jan E. and Yuan, Yiwen and Zhang, Zecheng and He, Xinwei and Leskovec, Jure},
	year = {2024},
	note = {arXiv:2407.20060 [cs.LG]},
}


%% file: references_manual.bib
@article{purucker2026beyond,
  title={Beyond IID: How General Are Tabular Foundation Models, Really?},
  author={Purucker, Lennart and Tschalzev, Andrej and Erickson, Nick and Blayer, Gioia and Holzm{\"u}ller, David and Arazi, Alan and Pfefferle, Alexander and Tajjar, Mustafa and Varoquaux, Ga{\"e}l and Hutter, Frank},
  journal={arXiv preprint arXiv:2606.30410},
  year={2026}
}

@article{arazi2026tabstar,
  title={{TabSTAR}: A tabular foundation model for tabular data with text fields},
  author={Arazi, Alan and Shapira, Eilam and Reichart, Roi},
  journal={Advances in Neural Information Processing Systems},
  volume={38},
  pages={172108--172161},
  year={2025}
}

@article{arazi2026multabench,
  title={{MulTaBench}: Benchmarking Multimodal Tabular Learning with Text and Image},
  author={Arazi, Alan and Shapira, Eilam and Grunblat, Shoham and Ventura, Mor and Hoffer, Elad and Blayer, Gioia and Holzm{\"u}ller, David and Purucker, Lennart and Varoquaux, Ga{\"e}l and Hutter, Frank and others},
  journal={arXiv preprint arXiv:2605.10616},
  year={2026}
}

@article{blayer2026strable,
  title={{STRABLE}: Benchmarking Tabular Machine Learning with Strings},
  author={Blayer, Gioia and Kim, Myung Jun and Lefebvre, F{\'e}lix and Purucker, Lennart and Arazi, Alan and Shapira, Eilam and Reichart, Roi and Hutter, Frank and Morvan, Marine Le and Holzm{\"u}ller, David and others},
  journal={arXiv preprint arXiv:2605.12292},
  year={2026}
}

@article{akhter2026fair,
  title={A Fair Benchmarking of Deep Relational Database Learning Models},
  author={Akhter, Kazi F and Ajendla, Bharath and Samad, Manar D},
  journal={arXiv preprint arXiv:2607.03659},
  year={2026}
}

@misc{relgnntrainingcodeissue,
	title = {Request for releasing training code},
	author = {{hydrogenhy}},
	year = {2025},
	howpublished = {GitHub issue \#3 on \texttt{snap-stanford/RelGNN}},
	url = {https://github.com/snap-stanford/RelGNN/issues/3},
}

@misc{relgnntuningissue,
	title = {Additional details regarding the tuning regime},
	author = {Hayler, Adrian},
	year = {2026},
	howpublished = {GitHub issue \#4 on \texttt{snap-stanford/RelGNN}},
	url = {https://github.com/snap-stanford/RelGNN/issues/4},
}

@misc{kumorfmhparamsissue,
	title = {Source of {RelBench} benchmarking hparams (and recent changes)? {Found} on validation or test data?},
	author = {Purucker, Lennart},
	year = {2026},
	howpublished = {GitHub issue \#74 on \texttt{kumo-ai/kumo-rfm}},
	url = {https://github.com/kumo-ai/kumo-rfm/issues/74},
}

@inproceedings{featuretools,
	author = {Kanter, James Max and Veeramachaneni, Kalyan},
	title = {Deep Feature Synthesis: Towards Automating Data Science Endeavors},
	booktitle = {2015 {IEEE} International Conference on Data Science and Advanced Analytics ({DSAA})},
	year = {2015},
	pages = {1--10},
	doi = {10.1109/DSAA.2015.7344858},
}

@misc{onebm,
	author = {Lam, Hoang Thanh and Thiebaut, Johann-Michael and Sinn, Mathieu and Chen, Bei and Mai, Tiep and Alkan, Oznur},
	title = {One Button Machine for Automating Feature Engineering in Relational Databases},
	year = {2017},
	url = {https://arxiv.org/abs/1706.00327},
	note = {arXiv:1706.00327 [cs.DB]},
}

@misc{getml,
	author = {{getML}},
	title = {{getML} Community Edition: Feature Learning for Relational Data and Time Series},
	howpublished = {\url{https://github.com/getml/getml-community}},
	year = {2022},
	note = {GitHub repository; includes the {FastProp} propositionalization algorithm. Accessed 2026-08-02},
}

@misc{featuretoolsrepo,
	author = {{Alteryx, Inc.}},
	title = {Featuretools: An Open Source Python Library for Automated Feature Engineering},
	howpublished = {\url{https://github.com/alteryx/featuretools}},
	year = {2017},
	note = {GitHub repository, created September 2017; 7,666 stars and approx.\ 197,000 monthly {PyPI} downloads as of 2026-08-02},
}

@misc{gnnpitfalls,
	title = {Pitfalls of {Graph} {Neural} {Network} Evaluation},
	url = {http://arxiv.org/abs/1811.05868},
	doi = {10.48550/arXiv.1811.05868},
	publisher = {arXiv},
	author = {Shchur, Oleksandr and Mumme, Maximilian and Bojchevski, Aleksandar and G{\"u}nnemann, Stephan},
	year = {2018},
	note = {arXiv:1811.05868 [cs.LG]},
}

@inproceedings{welltunednets,
	title = {Well-tuned Simple Nets Excel on Tabular Datasets},
	booktitle = {Advances in Neural Information Processing Systems ({NeurIPS})},
	url = {https://proceedings.neurips.cc/paper_files/paper/2021/hash/c902b497eb972281fb5b4e206db38ee6-Abstract.html},
	author = {Kadra, Arlind and Lindauer, Marius and Hutter, Frank and Grabocka, Josif},
	year = {2021},
}

@inproceedings{rlmatters,
	title = {Deep Reinforcement Learning that Matters},
	booktitle = {Proceedings of the {AAAI} Conference on Artificial Intelligence},
	volume = {32},
	doi = {10.1609/aaai.v32i1.11694},
	author = {Henderson, Peter and Islam, Riashat and Bachman, Philip and Pineau, Joelle and Precup, Doina and Meger, David},
	year = {2018},
}

@article{hposurvey,
	title = {Hyperparameter optimization: Foundations, algorithms, best practices, and open challenges},
	journal = {{WIREs} Data Mining and Knowledge Discovery},
	volume = {13},
	number = {2},
	pages = {e1484},
	doi = {10.1002/widm.1484},
	author = {Bischl, Bernd and Binder, Martin and Lang, Michel and Pielok, Tobias and Richter, Jakob and Coors, Stefan and Thomas, Janek and Ullmann, Theresa and Becker, Marc and Boulesteix, Anne-Laure and Deng, Difan and Lindauer, Marius},
	year = {2023},
}

@misc{kumorfmbenchmarkscripts,
	title = {{KumoRFM} benchmark scripts},
	author = {{Kumo.AI}},
	year = {2026},
	howpublished = {\texttt{benchmarks/} in \texttt{kumo-ai/kumo-rfm}, commit \texttt{e6c8ad5}},
	url = {https://github.com/kumo-ai/kumo-rfm},
}

@misc{kumorfmv1,
	title = {{KumoRFM}: {A} {Foundation} {Model} for {In-Context} {Learning} on {Relational} {Data}},
	url = {http://web.archive.org/web/20260702201348/https://kumo.ai/research/kumo_relational_foundation_model.pdf},
	publisher = {Kumo.AI},
	author = {Fey, Matthias and Kocijan, Vid and Lopez, Federico and Lenssen, Jan Eric and Leskovec, Jure},
	year = {2025},
	month = may,
	note = {Company whitepaper; archived copy, the original URL (\url{https://kumo.ai/research/kumo_relational_foundation_model.pdf}) is no longer online},
}

@inproceedings{rdblearncompanion,
	title = {No Need to Train Your {RDB} Foundation Model},
	author = {Xu, Linjie and Zhang, Yanlin and Gan, Quan and Wang, Minjie and Wipf, David},
	booktitle = {International Conference on Machine Learning ({ICML})},
	year = {2026},
	doi = {10.48550/arXiv.2602.13697},
	url = {http://arxiv.org/abs/2602.13697},
}

@article{scikit_learn,
      title={Scikit-learn: Machine Learning in Python},
      author={Fabian Pedregosa and Gaël Varoquaux and Alexandre Gramfort and Vincent Michel and Bertrand Thirion and Olivier Grisel and Mathieu Blondel and Andreas Müller and Joel Nothman and Gilles Louppe and Peter Prettenhofer and Ron Weiss and Vincent Dubourg and Jake Vanderplas and Alexandre Passos and David Cournapeau and Matthieu Brucher and Matthieu Perrot and Édouard Duchesnay},
      journal={Journal of Machine Learning Research},
      volume={12},
      pages={2825--2830},
      year={2011},
      eprint={1201.0490},
      archivePrefix={arXiv},
      primaryClass={cs.LG},
      url={https://arxiv.org/abs/1201.0490},
}

@misc{tabpfn_3_model_report,
    title = {{TabPFN}-3: {Technical} {Report}},
    shorttitle = {{TabPFN}-3},
    url = {http://arxiv.org/abs/2605.13986},
    doi = {10.48550/arXiv.2605.13986},
    urldate = {2026-08-03},
    publisher = {arXiv},
    author = {Grinsztajn, Léo and Flöge, Klemens and Key, Oscar and Birkel, Felix and Jund, Philipp and Roof, Brendan and Manium, Mihir and Hoo, Shi Bin and Bühler, Magnus and Garg, Anurag and Safaric, Dominik and Robertson, Jake and Jäger, Benjamin and Alessi, Simone and Hayler, Adrian and Moroshan, Vladyslav and Purucker, Lennart and Singer, Philipp and Arazi, Alan and Siems, Julien and Metzen, Jan Hendrik and Grab, Georg and Erickson, Nick and Guo, Siyuan and Kalfon, Eliott and Bing, Simon and Salinas, David and Cornu, Clara and Wehrhahn, Lilly Charlotte and Kriuchkova, Diana and Kaya, Kursat and Sidhoum, Lydia and Salmon, Marie and Chen, Jerry and Hulsebos, Madelon and LeCun, Yann and Müller, Samuel and Schölkopf, Bernhard and Gambhir, Sauraj and Hollmann, Noah and Hutter, Frank},
    month = may,
    year = {2026},
    note = {arXiv:2605.13986 [cs.LG]},
}

@inproceedings{transformers,
	title = {Transformers: State-of-the-Art Natural Language Processing},
	booktitle = {Proceedings of the 2020 Conference on Empirical Methods in Natural Language Processing: System Demonstrations},
	pages = {38--45},
	publisher = {Association for Computational Linguistics},
	url = {https://aclanthology.org/2020.emnlp-demos.6/},
	doi = {10.18653/v1/2020.emnlp-demos.6},
	author = {Wolf, Thomas and Debut, Lysandre and Sanh, Victor and Chaumond, Julien and Delangue, Clement and Moi, Anthony and Cistac, Pierric and Rault, Tim and Louf, Remi and Funtowicz, Morgan and Davison, Joe and Shleifer, Sam and von Platen, Patrick and Ma, Clara and Jernite, Yacine and Plu, Julien and Xu, Canwen and Le Scao, Teven and Gugger, Sylvain and Drame, Mariama and Lhoest, Quentin and Rush, Alexander},
	year = {2020},
}

@misc{paramfreeviable,
	title = {Parameter-{Free} {Encoders} {Remain} {Viable} for {RDB} {Foundation} {Models}},
	url = {http://arxiv.org/abs/2607.05476},
	author = {Xu, Linjie and Wipf, David},
	year = {2026},
	note = {arXiv:2607.05476. ICML 2026 Workshop on Foundation Models for Structured Data},
}

@article{demvsar2006statistical,
  title={Statistical comparisons of classifiers over multiple data sets},
  author={Dem{\v{s}}ar, Janez},
  journal={Journal of Machine learning research},
  volume={7},
  number={Jan},
  pages={1--30},
  year={2006}
}

@inproceedings{chiang2024chatbot,
  title={Chatbot arena: An open platform for evaluating llms by human preference},
  author={Chiang, Wei-Lin and Zheng, Lianmin and Sheng, Ying and Angelopoulos, Anastasios Nikolas and Li, Tianle and Li, Dacheng and Zhang, Hao and Zhu, Banghua and Jordan, Michael and Gonzalez, Joseph E and others},
  booktitle={International Conference on Machine Learning ({ICML})},
  series={Proceedings of Machine Learning Research},
  volume={235},
  pages={8359--8388},
  publisher={PMLR},
  year={2024}
}

@article{bradleyterry1952,
  title={Rank Analysis of Incomplete Block Designs: I. The Method of Paired Comparisons},
  author={Bradley, Ralph Allan and Terry, Milton E.},
  journal={Biometrika},
  volume={39},
  number={3--4},
  pages={324--345},
  year={1952},
  doi={10.1093/biomet/39.3-4.324}
}

@inproceedings{ni2019justifying,
	title = {Justifying {Recommendations} using {Distantly-Labeled} {Reviews} and {Fine-Grained} {Aspects}},
	author = {Ni, Jianmo and Li, Jiacheng and McAuley, Julian},
	booktitle = {Proceedings of the 2019 Conference on Empirical Methods in Natural Language Processing and the 9th International Joint Conference on Natural Language Processing (EMNLP-IJCNLP)},
	pages = {188--197},
	year = {2019},
	doi = {10.18653/v1/D19-1018},
}

@misc{amazonreviewdump,
	title = {Amazon {Review} {Data} (2018)},
	author = {Ni, Jianmo and Li, Jiacheng and McAuley, Julian},
	howpublished = {Dataset page, University of California San Diego},
	url = {https://cseweb.ucsd.edu/~jmcauley/datasets/amazon_v2/},
	year = {2018},
	note = {No license stated; the page requests that Ni et al. (2019) be cited},
}

@misc{kaggleavito,
	key = {Kaggle},
	title = {Avito {Context} {Ad} {Clicks}},
	howpublished = {Kaggle competition},
	url = {https://www.kaggle.com/competitions/avito-context-ad-clicks},
	year = {2015},
	note = {Competition rules require acceptance before download and are not publicly readable},
}

@misc{kagglehm,
	key = {Kaggle},
	title = {H\&{M} {Personalized} {Fashion} {Recommendations}},
	howpublished = {Kaggle competition},
	url = {https://www.kaggle.com/competitions/h-and-m-personalized-fashion-recommendations},
	year = {2022},
	note = {Competition rules require acceptance before download and are not publicly readable},
}

@misc{kaggleevent,
	key = {Kaggle},
	title = {Event {Recommendation} {Engine} {Challenge}},
	howpublished = {Kaggle competition},
	url = {https://www.kaggle.com/competitions/event-recommendation-engine-challenge},
	year = {2013},
	note = {Competition rules require acceptance before download and are not publicly readable},
}

@misc{ergastterms,
	key = {Ergast},
	title = {Ergast {Developer} {API}: {Terms} \& {Conditions}},
	howpublished = {Ergast Developer API},
	url = {http://web.archive.org/web/20240130210613/http://ergast.com/mrd/terms/},
	year = {2024},
	note = {Archived 30 January 2024. The API was deprecated during 2024 and shut down at the end of that year; \url{ergast.com} no longer serves the API or these terms},
}

@misc{stackexchangedump,
	key = {Stack Exchange},
	title = {Stack {Exchange} {Data} {Dump}},
	howpublished = {The Internet Archive},
	url = {https://archive.org/download/stackexchange},
	year = {2023},
	note = {The \texttt{license.txt} accompanying the dump states that contributed content is ``cc-wiki'' licensed, linking to CC BY-SA 3.0, and sets out attribution requirements for republishing content},
}

@misc{clinicaltrialsterms,
	key = {ClinicalTrials.gov},
	title = {{ClinicalTrials.gov} {Terms} and {Conditions}},
	howpublished = {U.S. National Library of Medicine},
	url = {http://web.archive.org/web/20230612201546/https://clinicaltrials.gov/ct2/about-site/terms-conditions},
	year = {2023},
	note = {Archived 12 June 2023; the current page is rendered client-side and is not archivable as text},
}

@misc{relbench_leakage_fix,
	author = {{The RelBench maintainers}},
	title = {Fix rel-event (user-ignore) and rel-stack (db), fix task names},
	howpublished = {\url{https://github.com/snap-stanford/relbench/pull/357}},
	year = {2026},
	note = {GitHub pull request \#357 on snap-stanford/relbench, merged 2026-02-01; regenerates the rel-event/user-ignore task table (temporal-leakage fix) and the rel-stack database; first contained in relbench 2.1.0},
}
